\documentclass[prb]{revtex4-2}

\usepackage{color}

\usepackage{hyperref}
\usepackage[noend]{algpseudocode}
\usepackage{algorithm}

\definecolor{eggplant}{RGB}{126,93,181}
\definecolor{cayenne}{RGB}{148,17,0}
\definecolor{teal}{RGB}{0,145,147}
\definecolor{blueberry}{RGB}{4,51,255}

\usepackage{graphicx}

\usepackage{amsmath}     
  \usepackage{cleveref}	
  	\crefname{figure}{Fig.}{Figs.}
  	\crefname{table}{Table}{Tables}
  	\crefname{equation}{Eq.}{Eqs.}
  	\crefname{section}{Section}{Sections}
  	\crefname{subsection}{Section}{Sections}
  	\crefname{subsubsection}{Section}{Sections}
  	\crefname{algorithm}{Algorithm}{Algorithms}

  \usepackage{amssymb}
  \usepackage{bm}
  \usepackage{mathrsfs}
  \usepackage{titlesec}
  \usepackage{subcaption}

\date{\today}

\begin{abstract}
    In this work we present a method to accelerate the optimization of learning high dimensional functions using deep neural network (DNN). 
    This optimization procedure introduces contextual features into the first layer of a DNN.
    The parameters of DNN are optimized via standard gradient descent while keeping the input-feature basis fixed.
    After optimization of the DNN parameters, the feature layer is provided a chance to update and change before DNN optimization resumes.
    The feature layer has two types of functions: those that can be evaluated quickly in a matrix-free way on the domain (i.e. rank-1 features) and more complex features that must first be decomposed using tensor network (TN) decomposition strategies (tensor features).
    In particular, we study the effect of adding features which distill pretrained DNN into TNs using a discretize and decompose strategy.
    To efficiently decompose high-dimensional functions constructed from discretized DNN, we leverage a randomized tensor decomposition strategy.
    Using randomization, we are able to reduce the storage cost of decomposing high dimensional functions by at least 8 orders of magnitude.
    Using this approach, we are able to efficiently train models between 5 and 40 dimensions.
\end{abstract}
\begin{document}

\author{Karl Pierce$^\dagger$}
\author{Yuehaw Khoo$^*$}
\author{Haizhao Yang$^\dagger$}
\affiliation{$^{\dagger}$ The University of Maryland, College Park, College Park MD 20742 USA}
\affiliation{$^{*}$ The University of Chicago, Chicago, IL 60637 USA}

\title{Accelerated Learning of High Dimensional Functions with a Tensor-Featured Training Network}

\maketitle
\section{Introduction}
Methods to interpolate continuous functions and solve partial differential equations (PDEs)
computationally are an interesting and on-going challenge in scientific computing. 
Canonical methods to represent continuous functions have require users to construct nontrivial grid-based meshes or introduce complicated numerical techniques.
Not only are these strategies complicated from an implementation standpoint, but they can also require large amounts of computational resources. 
Grid-based strategies, in particular, suffer from the curse of dimensionality which states that the cost of storing and accessing elements of a multidimensional array grows exponentially with the number of dimensions in the array.
As a means to break the curse of dimensionality and solve large and high-dimensional problems
scientific machine learning (ML) has recently emerged.\cite{Jordan2015,Raissi:2019:JCP}

Scientific ML is a powerful set of tools which may be used to compactly represent high-dimensional and unknown functions.
These tools work by learning the underlying structure of functions using a non-linear network of tuneable parameters.
Because the scientific ML models do not attempt to memorize values of a function at specific points on a mesh, these models are commonly called mesh-free methods.
By construction, these models are extremely expressive and the number of optimizable components in the models is not strongly correlated to the dimensionality of the function. 
Though scientific ML models are exceptionally good at 
representing a diverse range of functions, 
the optimization of a ML model can be an extremely inefficient process, requiring hundreds of compute hours even on today's high-throughput GPU processing units.
This is especially true for deep neural network (DNN) optimizations, 
like physics informed NNs (PINNs), as their optimization is non-convex.\cite{Hu2024,Urbn2025}
In general, training ML models become more difficult and slowly converging with increasing function dimensionality.
There have been many ideas to accelerate the optimization process, some examples include sampling techniques to target high-residual sub-domains in the training process\cite{Gu2021SelectNet:Equations} and methods to construct random features using the kernel method.\cite{Liao:2026:CNSNS}

In this work, we are interested in using ideas from tensor-networks (TN) methods to improve the efficiency of training DNNs.
TNs are becoming increasingly popular in scientific computing as methods to efficiently represent high-dimensional functions and data.\cite{Richter2021}
Effectively, tensor network methods seek to factorize high-dimensional functions into a sums of outer products of low-dimensional component functions.
Using this factorization technique, tensor network methods are able to break the curse of dimensionality.
Unlike the mesh-free ML models, TN methods do require a collocation grid.
As a result, while the exact storage complexity of the method depends on the topology of the TN, in general there is a strong correlation between cost of storing a TN, the dimension of the function being decomposed and number of collocation points per dimension.
How well a TN can represent a function explicitly depends on the type of function and the topology of the TN. 
Though these properties make TN in general more difficult to apply to problems than ML models, TN methods do shine in their ability to be efficiently optimized.
This is because most tensor-network algorithms rely on linear algebra techniques such as the singular value decomposition whereas machine learning models rely on non-convex gradient descent techniques.
Therefore, in this work we seek to determine methods to apply the beneficial properties of TN methods, i.e. reliable  optimization and interpretability, to improve and accelerate the optimization of DNN.

Our idea is similar to a warm-start initialization technique and works in the following way.
We generate a collection of functions and apply these functions to input data to a DNN.
We add the output of these functions to the input layer of the DNN and optimize parameters of the DNN subject to the input data and these new features.
A most straight-forward feature would be a previously optimized DNN. 
We demonstrate in this work, why this idea fails and instead use a TN decomposition strategy to improve on this ides.
Furthermore, we identify two distinct subclasses of basis functions, those that can be applied quickly in a matrix-free way and those that can be approximated via a TN decomposition.
We, therefore, introduce a 2-step optimization process for DNNs where, given a set of input features, DNN parameters are optimized recursively by canonical gradient descent techniques for a fixed number of iterations. 
After parameter optimization, the basis functions are evaluated and updated to improve convergence.
After the basis functions are modified, a new DNN is 
initialized and parameter training is resumed.
This work draws parallels with random feature training as the DNN now could be considered the combination of it's input layer and neuron layers. Like random feature training, we fix the parameters associated with input level features during DNN training.
One major difference is that the interaction of features is facilitated implicitly by the DNN parameter layers.
This gives the feature layer of the DNN an outer-product structure similar to a rank-1 TN decomposition.
Therefore, we call this technique {\bf Tensor-Featured} training.

As pointed out previously, we will introduce DNNs which are discretized and decomposed on their domain.
To do this decomposition we will utilize the canonical polyadic decomposition (CPD).\cite{hitchcock1927expression,carroll1970analysis,harshman1970foundations} 
We limit this work to a single tensor decomposition strategy to simplify the analysis of the procedure.
In practice any flavor or combination of tensor decompositions could be enabled.
Furthermore, one can contextualize the tensor decompositions, like the CPD and the tensor train decomposition, as method for computing a high dimensional Gaussian process using a kernel matrix which is constructed as a tensor product of one-dimensional kernel matrices.\cite{Wesel:2025,Yuan:2026:Arxiv}
As a result, we hope to demonstrate the following:
first that the introduction of tensor features acts as a regularizer in the optimization of DNN and; second that the tensor decomposition acts as a smoothing process for the DNN via non-local features introduced to the tensor kernel matrix.
Because tensor decompositions can be extremely expensive in both computer time and memory, we will leverage advancements in randomized linear algebra to efficiently decompose high dimensional functions.

The remainder of the text is laid out in the following manner.
In \cref{sec:Theory}, we will discuss the theoretical background of the work 
including the randomized canonical polyadic decomposition and the relationship between tensor decompositions and the gaussian process for high-dimensional functions.
In \cref{sec:experiments} we describe the outlines of our experiments to learn functions which are solutions to PDEs in various degrees of dimensionality spanning from $5$ to $40$.
In \cref{sec:res} we provide results for our optimization experiments, compare our result to canonical optimization strategies and discuss methods to improve the optimization process for DNN to learn unknown functions.
Finally in \cref{sec:conclusion} we summarize the findings and discuss future directions and improvements for this work.

\section{Theoretical Background}\label{sec:Theory}

    \subsection{The Canonical Polyadic Decomposition}
    The CPD is a higher-order extension of the singular value decomposition and represents a function of $d$ variables (expressed as an order-$d$ tensor) as a sum of outer products of single variables, i.e. given $\mathcal{T} \in \mathbb{R}^{I_1, I_2, \dots, I_d}$
    \begin{align}\label{eq:cpd}
        t_{a,b,\dots,n} \overset{\mathrm{CPD}}{=} \sum_{i=1}^{R_\mathrm{CP}} \lambda_i {\bf a}_i \circ {\bf b}_{i} \circ \dots \circ {\bf n}_i
    \end{align}
    where $\circ$ defines the vector outer product, ${\bf a}_i \in \mathbb{R}^{I_1}, {\bf b}_i\in \mathbb{R}^{I_1}, \dots, {\bf n}_i \in \mathbb{R}^{I_d}$, $\lambda_i$ is a scaling constant that allows the vectors $[a_i,\dots n_i]$ be unit normalized, and $R_{\mathrm{CP}}$ is the CP rank, i.e. the smallest set of vectors such that \cref{eq:cpd} satisfies the equality.
    All the vectors which span the same mode of the tensor can be group into a factor matrix, i.e. ${\bf A} = [ {\bf a}_1, \dots, {\bf a}_{R_\mathrm{CP}}]\in \mathbb{R}^{I_a \times R_\mathrm{CP}}$.
    Therefore, the CPD reduces the storage complexity of a $d$-dimensional function from $N^d$ where $N \approx I_1 \approx I_2,...$ to $N D R_{\mathrm{CP}}$.
    Although finding the exact CP rank is formally an \textit{NP}-hard problem, it is straightforward and easy to discover an approximate CPD of $\mathcal{T}$ given a desired rank $R$.
    By construction, the CPD defines a set of 1-dimensional basis functions such that a function of $d$ inputs as a separable set of products of functions of single inputs.
    
    The most standard way to optimize the rank-$R$ CPD is via the alternating least squares (ALS) optimization.\cite{VRG:kroonenberg:1980:P,VRG:beylkin:2002:PNAS}
    The ALS method splits the optimization of the CPD into $d$-linear least square problems.
    Unfortunately, the CPD-ALS optimization is non-convex.
    However, based on our experience and results, we find that the non-convex nature of the CPD-ALS has minimal effect on discovering accurate decomposition's of tensor for most cases in physics and chemistry.\cite{VRG:pierce:2021:JCTC}
    As we will demonstrate in \cref{sec:res}, this is also true for this work.
    A more serious issue with the CPD-ALS is that the optimization suffers from the curse of dimensionality, i.e. the computational cost of the method scales exponentially with the dimension of the function being decomposed.
    
    In general, the task of the CPD-ALS is to iteratively solve a collection of extremely overdetermined least squares problems.
    Because these problems are over-determined, a direct result of the matricization of higher order tensors along a single mode of the tensor, the CPD-ALS a perfect candidate for randomized linear algebra. 
    There are two main ways to apply randomization to the CPD-ALS, via sampled least squares\cite{battaglino2018practical,larsen2022practical} or via sketched and sampled least squares.\cite{Woodruff:2014:FTCS,Rokhlin:2008:PNAS}
    In this work, we will only consider the sampled least squares method because of its low-cost and structure preserving properties.
    In the literature, it has been demonstrated that the number of samples required for the CPD randomized ALS (CPD-RALS) is proportional to the rank of the CPD and has no direct dependence on the dimensionality of the function being decomposed.\cite{battaglino2018practical,Fakih2026AcceleratingDecomposition}
    Therefore, in this work, we will assume that the CP rank of every tensor is practically small compared to the number of interpolation points along a single dimension and does not grow (significantly) with increasing tensor dimension.
    The RALS procedure therefore reduces the computational complexity of the CPD from exponential in function dimension to polynomial and independent of function dimension.
    Using the CPD-RALS, the most time intensive component of the algorithm is the cost of sampling the target function which grows as $dSN$ where $d$ is the dimension of the function, $S$ is the number of samples per mode and $N$ is the number of interpolation points along a single dimension.
    To minimize the overall cost of this step, we will assume that single set of samples gathered from a uniform distribution is sufficient for optimizing the CPD-RALS. 
    
    In this work we will not explicitly derive the sampled CPD-RALS method. 
    We utilize a modified version of the canonical version developed by Battaglino et. al. where samples are determined via the leverage score distribution of each factor matrix.
    Our method deviates from the canonical method in the following way.
    We found that the magnitude of elements in the sampled Khatri-Rao product (KRP) associated CPD-RALS decay's rapidly with dimensionality of the tensor.
    To accommodate this issue, we introduce an intermediate normalization where rows of the KRP are normalized to 1 such that
    \begin{align}\label{eq:norm}
        Ax &= B \\ \nonumber
        (l\hat{A})(xl^{-1}) &= B \\ \nonumber
        \hat{A}\bar{x} &= B.
    \end{align}
    The intermediate normalization tensor $l$ can, without loss of generality, be factored into the column-wise normalization of $\bar{x}$ and, therefore, absorbed into $\lambda$ in \cref{eq:cpd}.
    We have found that this process, in general, stabilizes the LS solver in the CPD-RALS procedure.
    
    \subsection{The CPD-ALS as Gaussian Processes}
    Recently, it has been shown that there is a connection between popular tensor decomposition strategies, kernel machines\cite{NIPS2016_5314b967} and Gaussian process(GP).\cite{Wesel2024}
    The GP is a mechanism for modeling functions and allows the user to incorporate prior knowledge of the function (i.e. random inputs on the domain and associated noisy observations).
    GP are fully described by two components, a mean $\mu \in \mathbb{R}$ which is usually chosen to be zero and a covariance or kernel matrix $k \in \mathbb{R}^{\Omega_i \times \Omega_i}$ where $\Omega_i$ is the extent of the domain in the $i$th dimension.
    Given a set of points and their associated observations, one can determine random functions on the domain by sampling the multivariate gaussian associated with this mean and kernel matrix.
    The kernel matrix is commonly approximated using a finite number of basis functions\cite{Candela:2005,Rasmussen2006}
    \begin{align}
        k(x,x') = \varphi(x)^T \Lambda \varphi(x')
    \end{align}
    where $\varphi(x) \in \mathbb{R}^{\Omega_i\times M}$ are basis functions and $\Lambda \in \mathbb{R}^{M \times M}$ are basis function weights.
    One can recognize that each CPD factor matrix are explicitly constructed as a collection of $R$, 1-dimensional basis functions for a on set of observed points on a predefined collocation grid.
    
    The CDP-ALS iteratively constructs each 1-dimensional basis matrix (CPD factor matrix) in the following way. 
    For convenience we limit this explanation to a three dimensional function $\mathcal{T} \in \mathbb{R}^{I_a \times I_b \times I_c}$ but this idea can be extended to arbitrary dimension without loss of generality.
    First, the loss function is defined for the $I_a$th mode
    \begin{align}
        f(A) = \|T_{a} - A^* [B \otimes C]^T \|^2_2
    \end{align}
    where $T_{a}$ is the so called matriciation of the third order tensor $\mathcal{T}$ that permutes and reshapes the tensor to be $T_{a} \in \mathbb{R}^{I_{a} \times I_{b}I_{c}}$, a short and long matrix.
    Also $\otimes$ defines the Khatri-Rao tensor product (KRP) which is the column-wise Kronecker product of two tensors.
    The KRP defines the following product $W_r = b_r \circ c_r$ where is a matrix $W_{r} \in \mathbb{R}^{I_b \times I_c}$ for each value of $r \in \mathbb{R}^R$.
    By definition $B$ and $C$ are also collections of 1D basis which were either chosen at random or drawn randomly a different GP.
    Next, one solves for $A^*$ by taking the gradient of the loss function and solving the following least squares problem
    \begin{align}\label{eq:normal}
        T_{a} (B \otimes C) = A^* (B \otimes C)^T (B \otimes C).
    \end{align}
    This equation consists of two parts: $T_{a} (B \otimes C) \in \mathbb{R}^{I_a \times R}$ which down-folds the high dimensional observations of the function into a single dimensional set of mean values for each basis basis function by taking the inner product of the higher order tensor and all other degrees of freedom.
    The second part is the construction of the squared KRP, $(B \otimes C)^T (B \otimes C)$.
    Using Tensor Algebra, this expression can be rewritten as $(B^T B) * (C^T C)$ where $*$ defines the Hadamard or element-wise product of two matrices.
    Written in this way one can see that $B^T B \in \mathbb{R}^{R \times R}$ and $C^T C \in \mathbb{R}^{R \times R}$ are both approximations to 1-dimensional kernel matrices.
    Therefore the squared KRP can be viewed as a product kernel\cite{Rasmussen2006} of many 1-dimensional kernel matrices.
    As a result, the above solution to the CPD-ALS can be viewed as a method that leverages GP to iteratively draw and refine 1-dimensional basis functions for high dimensional functions.
    By leveraging the basis function approximation and kernel product structure, the CPD-ALS is able to draw GP basis functions with information across the entire domain of the function with a relatively low cost. Unfortunately, one can see that the construction of the left hand side of \cref{eq:normal} suffers from the curse of dimensionality.

    In an effort to minimize the cost of the CPD-ALS, the CPD-RALS modifies the above process. 
    The gradient of the CPD-RALS loss function for the $I_a$th mode can be written as
    \begin{align}
        (T_{a} S_{a})[S^T_a(B \otimes C)] = A^* [S^T_{a} (B\otimes C)]^T[S^T_{a} (B\otimes C)]
    \end{align}
    where $S_{a} \in \mathbb{R}^{I_{b} I_{c} \times I_{s}}$ is a sparse, column-wise selection matrix chosen specifically for the $I_a$th subproblem.
    By constructing the randomized solver in this way we may significantly reduce the cost of inner-product down-folding process from $I_{a}I_{b}I_{c}R$ to $I_{a}I_{s}R$, provided that $I_{s} \propto R$ and $R \ll I_{b} I_{c}$.\cite{battaglino2018practical,Fakih2026AcceleratingDecomposition}
    However, this sampling process changes the form of our kernel matrix
    The sampling matrix first down-folds the high-dimensional space orthogonal to the $I_a$ into a relatively small set of observations.
    In this work, we choose these observations by
    first constructing a leverage score distribution of the kernel basis function.
    By taking advantage of the KRP structure of the kernel basis functions, we may randomly sample the leverage score distribution of each basis matrix independently.
    The row-wise leverage score distribution of each factor matrix measures how far away observations on each quadrature point are from all the other observations and thus the leverage scores are correlated with the degree of misprediction of a least squares model.
    Therefore, by importance sampling the KRP using the leverage score distribution, we are attempting to determine a compact basis of observations which contribute most strongly to the basis-approximated kernel matrix.
    After choosing this compact set of basis points, we now have a single set of kernel basis functions.
    In this process you can recognize that our normalization process in \cref{eq:norm} is effectively a modification of our basis function weights, $\Lambda$.

\section{Computational Experiments}\label{sec:experiments}
As this work serves as a preliminary study into this improved optimization process, we make a number of restrictions to simplify and limit the analysis.
First we only consider functions of dimension, $d \in [5,40]$.
Unless otherwise stated, we will use the following parameters by default in this manuscript.
In all cases we consider a DNN to have 3 layers and with 100 features per layer.
We apply the following activation functions to each layer $[\mathrm{relu}, \mathrm{relu}, \mathrm{identity}]$ of the DNN.
To optimize the parameters of the DNN we utilize an ADAM gradient descent algorithm.
We stop the parameter update of each DNN after 3000 epoch.
We consider one function that is time independent and one function which contain a time dependence.
For our test and training sets, we use a spatial domain of (-1,1) and a time domain of (0,1) and determine points on this domain using a regular grid of 10,000 points per mode.
The training set contains 1000 random points and the validation set contains 5000 random points.
For this study a single batch of training points is drawn for all DNN optimizations and this set is not updated.
This decision comes from an analysis of the structure of the randomized CPD-ALS optimization where we find no significant outliers in leverage scores across any single spatial dimension for all tested functions.
This decision was validated by decomposing each known functions via the CPD-RALS procedure using a single uniformly drawn random sample.
Resampling may become necessary in future works when considering more difficult learning settings such as increased dimensionality.

For the CPD, unless stated otherwise, we decompose DNN's on a grid of 500 points per mode and to a rank of $30$.
For the CPD-RALS procedure, we fix the number of samples to $3000$ per least squares subproblem.
This reduces the size of target tensor in each least squares subproblem from $500 \times 500^{d-1}$ to $500 \times 3000$. 
In the smallest tested case, this is a 20 million times reduction in storage cost of the CPD optimization process.
The CPD was optimized using 100 ALS iterations; however, early stopping conditions could be successfully applied to reduce the cost of the decomposition.
When necessary, to evaluate points which are not explicitly on the quadrature grid of the CPD, a simple nearest neighbor interpolation strategy is applied.

The experiments are primarily run on a Mac M1 Max equipped with 8 high-throughput CPU cores each with a 3.2GHz processing speed and 64 GB of unified RAM.
DNN were constructed and optimized using the Julia-based Flux.jl\cite{Flux.jl-2018,innes:2018} library.
CPD were constructed and optimized using the Julia-based ITensorCPD.jl library\cite{ITensorCPD.jl}

\section{Results and Discussion}\label{sec:res}

    \subsection{Iterative Tensor-Featured Training via Tensor Decomposed DNN.}
    First we investigate the effect of using a pretrained DNN and as a tensor feature.
    For this section we attempt to learn a function which is the solution to the the nonlinear elliptic equation inside a bounded domain,
    \begin{align}\label{eq:func1}
        u(x) = \sin(\frac{\pi}{2} (1-|x|_2)^{2.5}),
    \end{align}
    using the DNN structure described previously.
    We consider the problem in three dimensions $d=5, 10,$ and $15$.
    In this experiment, we first use canonical optimization strategies to train a DNN over a 3000 training epoch. 
    Throughout this work we will referred to this canonical training as \textbf{Conventional Training}.
    After this training we initialize two new DNN each with 1 additional feature added to their input layers. 
    For one DNN, this new feature is the output of the first DNN evaluated at the input training coordinates.
    We call this \textbf{NN Featured} training.
    In the other DNN, we provide the approximated output of the first DNN via CPD interpolated at the input training coordinates.
    As explained early, for off-grid points, we use a nearest neighbor interpolation. 
    We denote this second DNN as a \textbf{Tensor-Featured} trained model.
    
    \begin{figure}
        \begin{subfigure}{0.33\textwidth}
            \includegraphics[width=.9\columnwidth]{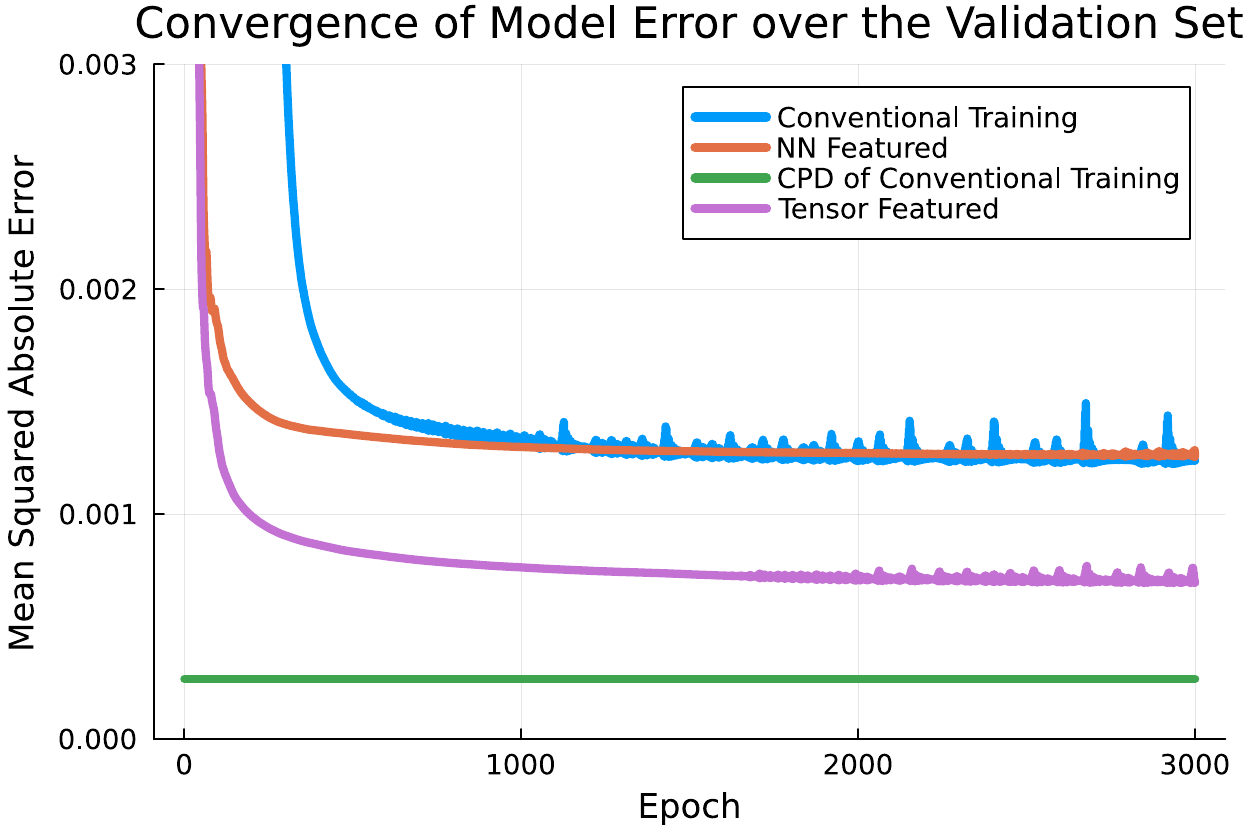}
            \caption{}
        \end{subfigure}\hfill
        \begin{subfigure}{0.32\textwidth}
            \includegraphics[width=.9\linewidth]{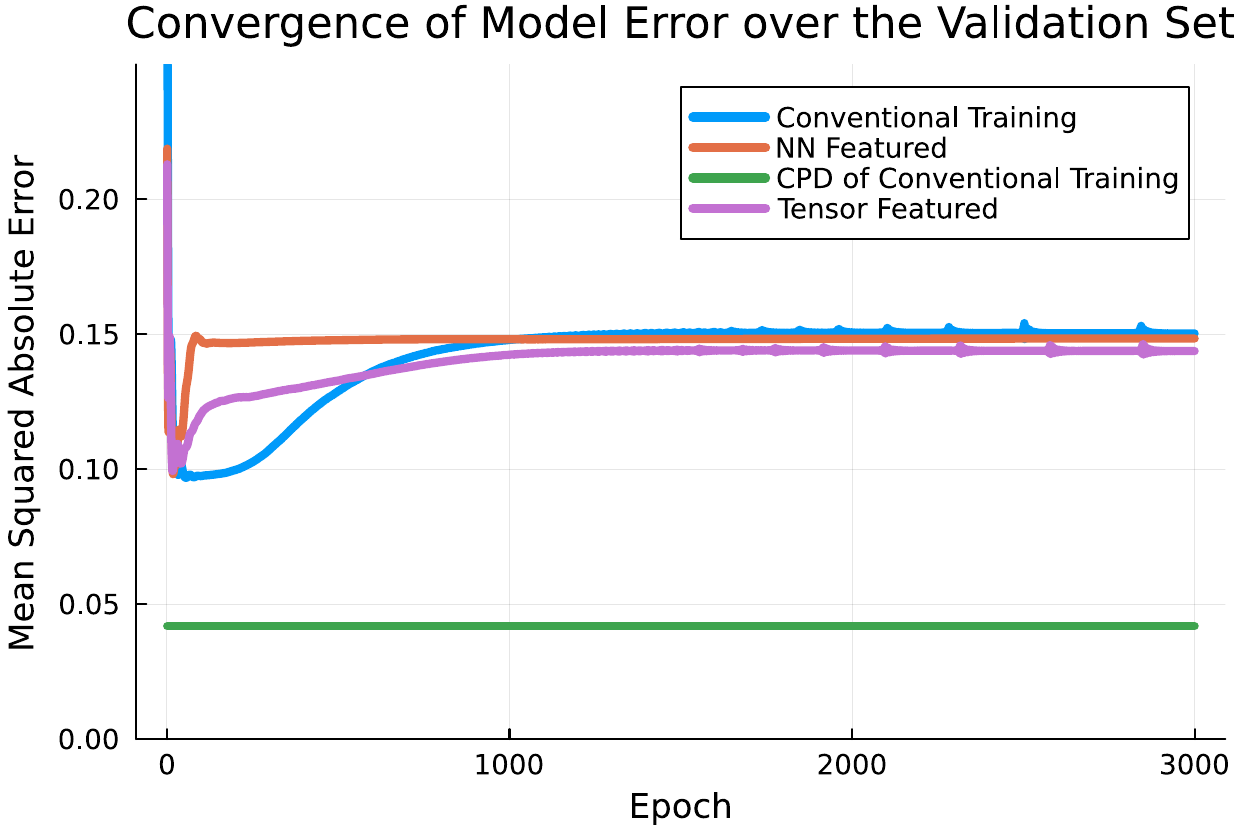}
            \caption{}
        \end{subfigure}\hfill
        \begin{subfigure}{0.32\textwidth}
            \includegraphics[width=.9\linewidth]{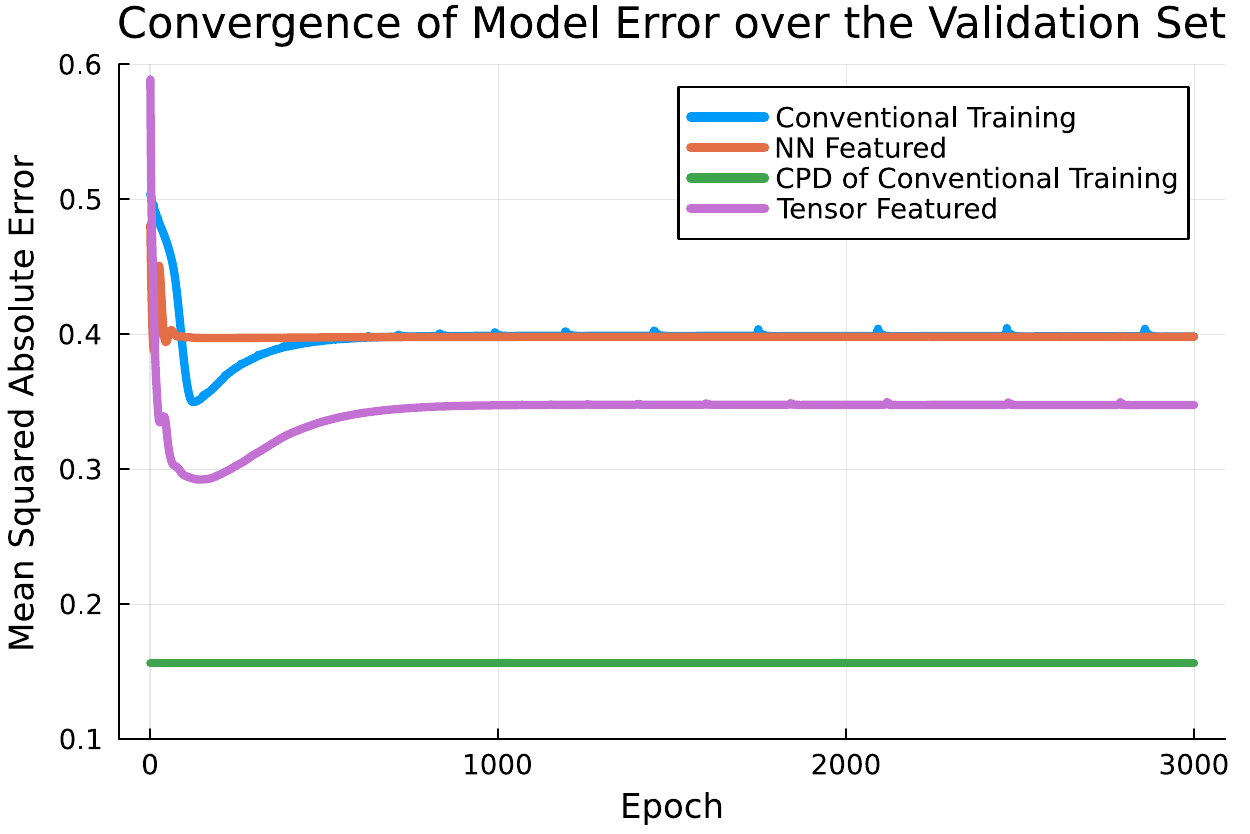}
            \caption{}
        \end{subfigure}\hfill
        \caption{Comparison of the convergence of different training procedures to learn \cref{eq:func1} in (a) $5$, (b) $10$, and (c) $15$ dimensions.}
        \label{fig:SFImprovment}
    \end{figure}
    
    In \cref{fig:SFImprovment} we compare accuracy of the three training strategies (conventional, NN featured and Tensor-Featured) on the set of validation points over 3000 gradient updates. 
    For reference, we also show the error of the CPD interpolation of the conventionally trained NN on the same set of validation points.
    One can see, across the board, decomposing the conventionally optimized DNN into a TN can significantly improve the accuracy of learned function. 
    As pointed out previously, we believe this is because the process of optimizing the tensor decomposition using non-local information which effectively smooths the DNN function over the domain.

    With respect to Tensor-Featured training, in \cref{fig:SFImprovment} one can also see that using the DNN directly as a feature does not practically provide any improvement over the conventional training.
    On the other hand with the CPD interpolation as a feature, we find an improvement in accuracy over the conventional training.
    Unfortunately, although there is an improvement in accuracy with the Tensor-Featured training in the validation set, the improvement is not dramatic and, in all cases, the CPD approximation alone outperforms the Tensor-Featured DNN.
    We believe this phenomena is not a failure in the Tensor-Featured learning but, instead, associated with the accuracy of the conventionally trained model relative to the ground truth.
    
\begin{figure}
        \begin{subfigure}{0.49\textwidth}
            \includegraphics[width=.8\columnwidth]{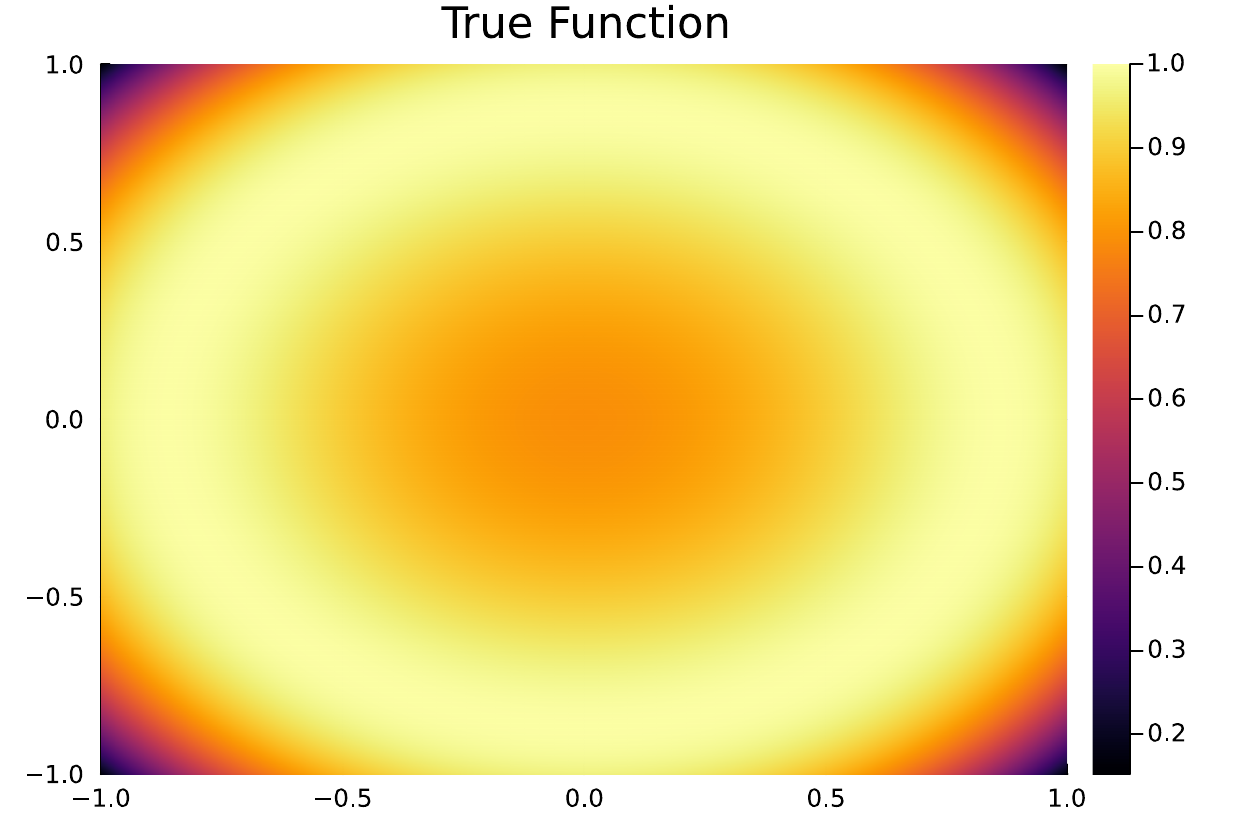}
            \caption{}
        \end{subfigure}\hfill
        \begin{subfigure}{0.49\textwidth}
            \includegraphics[width=.8\linewidth]{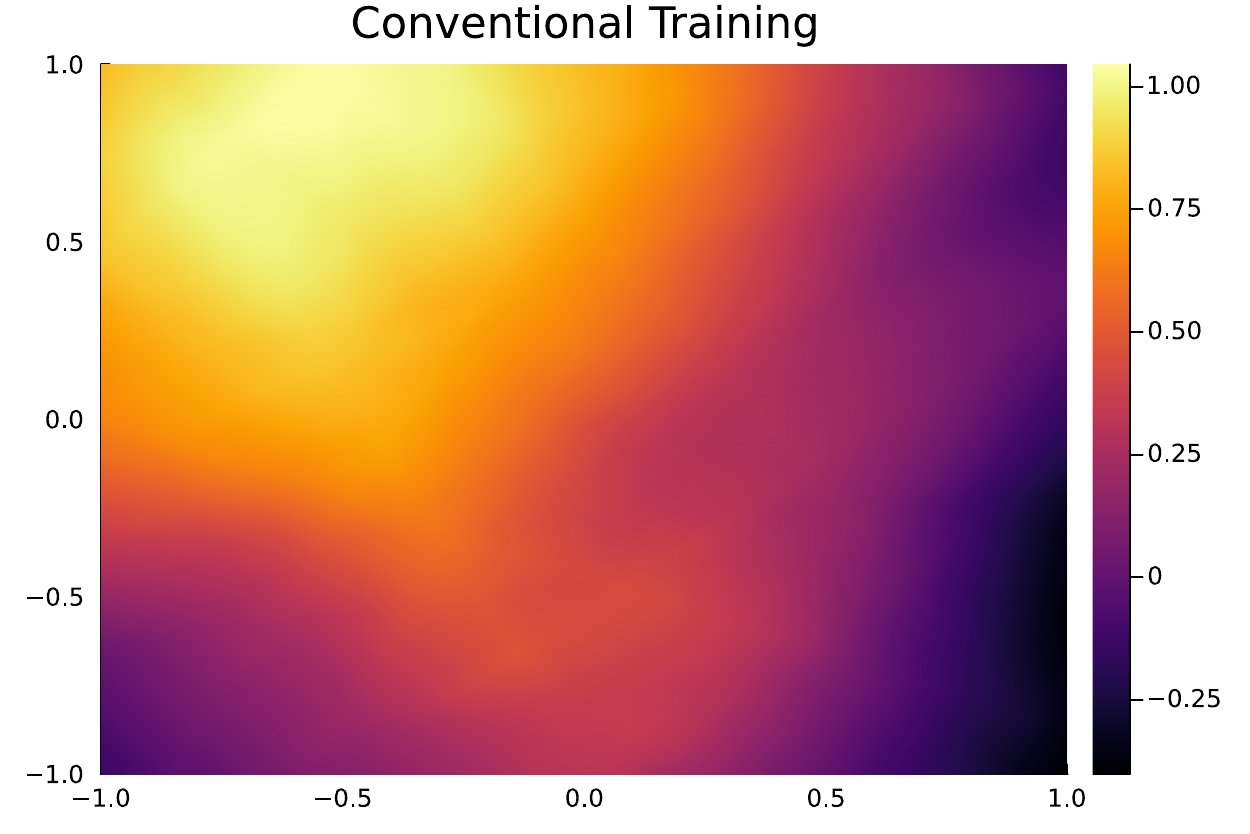}
            \caption{}
        \end{subfigure}
        \begin{subfigure}{0.49\textwidth}
            \includegraphics[width=.8\linewidth]{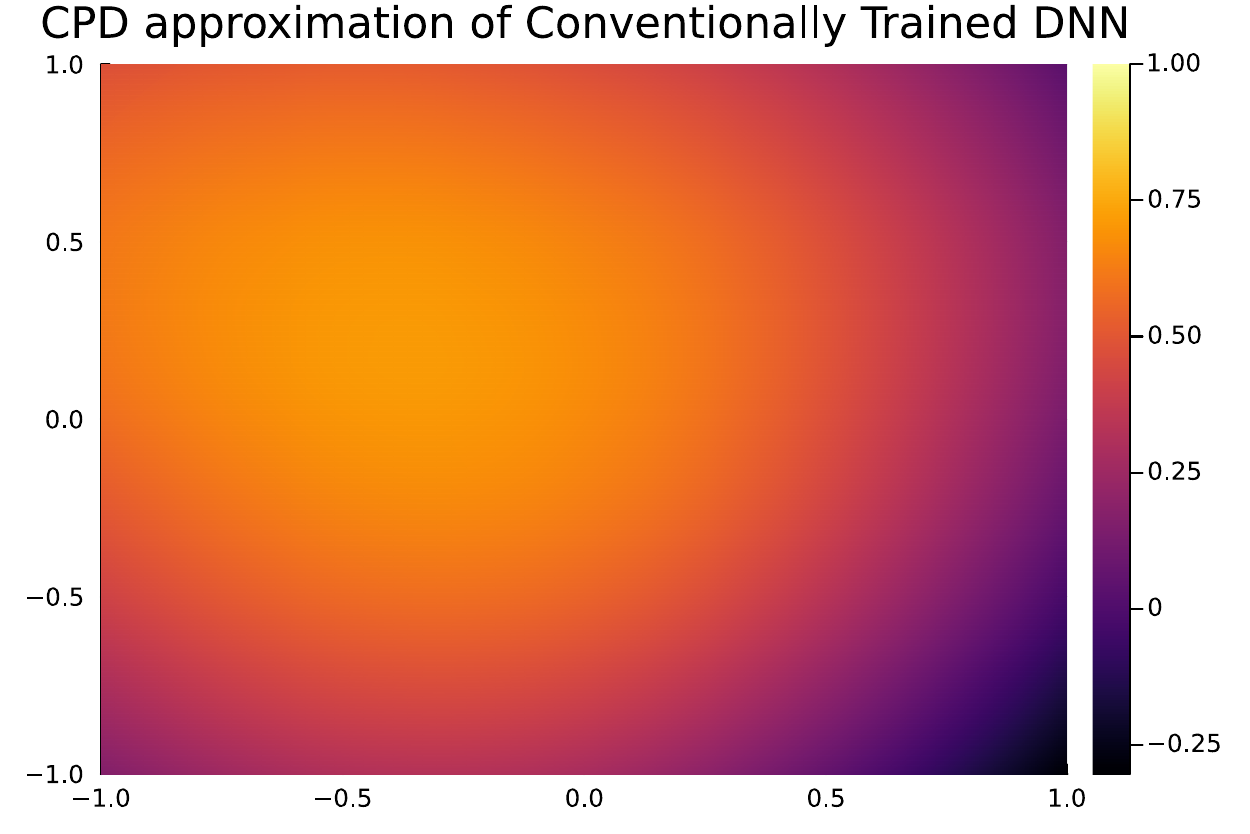}
            \caption{}
        \end{subfigure}\hfill
        \begin{subfigure}{0.49\textwidth}
            \includegraphics[width=.8\linewidth]{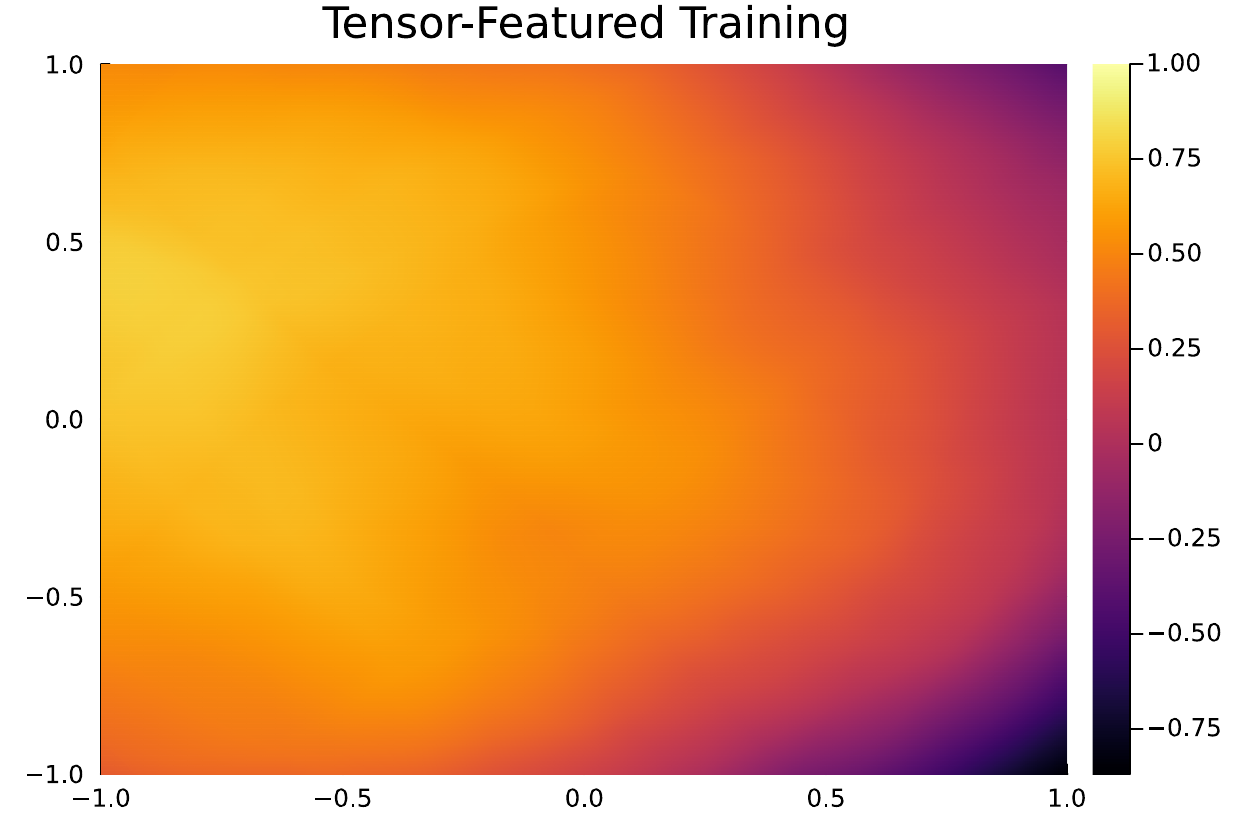}
            \caption{}
        \end{subfigure}\hfill
        \caption{Heatmaps of (a) the true function (\cref{eq:func1}), (b) the function learned by the conventionally trained DNN, (c) the CPD approximation of the conventionally trained DNN and (d) the Tensor-Featured trained DNN. Heatmaps are evaluated on a 2000 point interpolation grid in all modes and all but 2 modes are fixed to the 500th point on their interpolation grid. The DNN training is fixed to 1,000 samples in the training set.}
        \label{fig:heatmap1}
    \end{figure}

    To clarify this statement, consider the heatmaps in \cref{fig:heatmap1} which illustrates true and predicted values of \cref{eq:func1} on a 2000 point interpolation grids with all but 2 modes of the tensor fixed to the 500th interpolation point.
    One can see that, though \cref{fig:SFImprovment} implies that the CPD approximation is a better representation of the true function, none of these models look significantly more accurate in this representation.
    With this example, we demonstrate that because the reference function for the CPD (the conventionally trained network) is not representation of the ground truth, its addition to the feature list in the Tensor-Featured training doesn't significantly improve the DNN training.
    Therefore, we hypothesize that, if the conventional training is more similar of the ground truth, then the Tensor-Featured training procedure will be significantly accelerated.
    
    \begin{figure}
        \begin{subfigure}{0.49\textwidth}
            \includegraphics[width=.8\columnwidth]{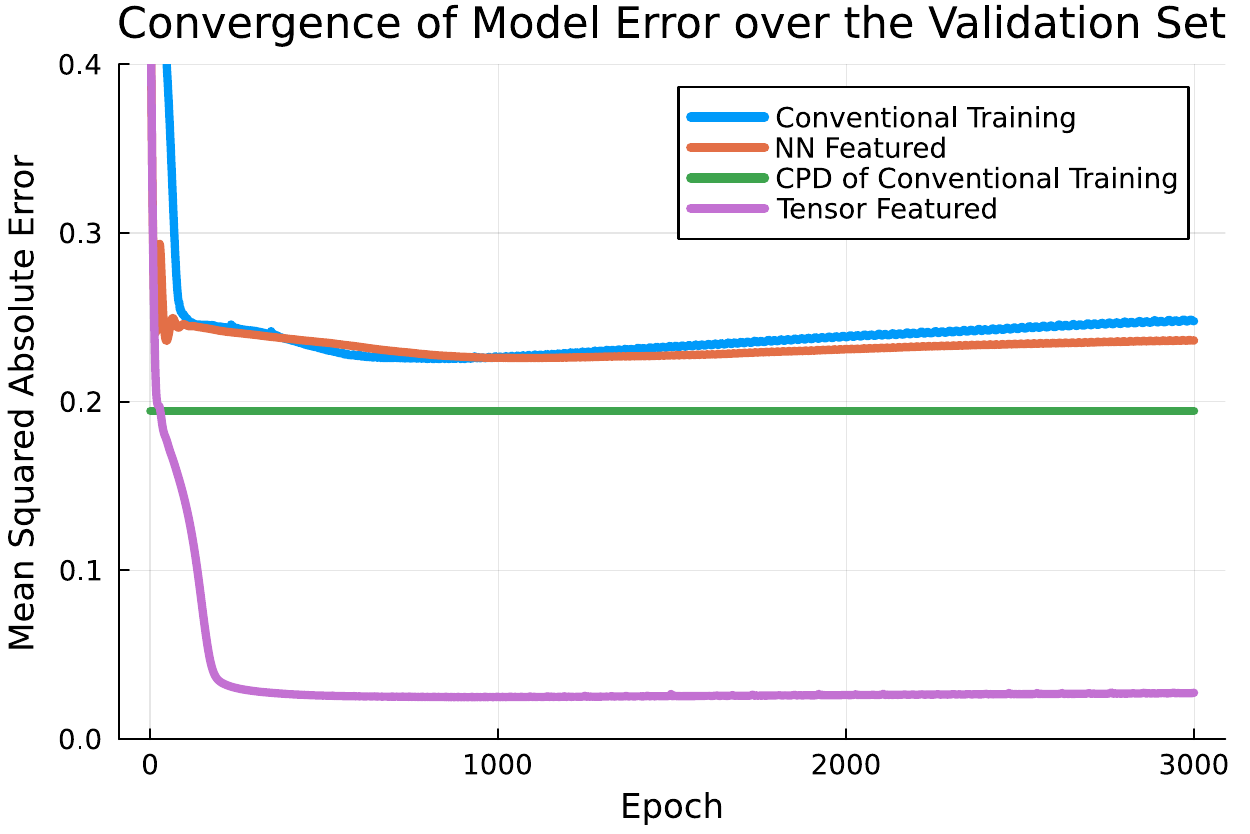}
            \caption{}
        \end{subfigure}\hfill
        \begin{subfigure}{0.49\textwidth}
            \includegraphics[width=.8\linewidth]{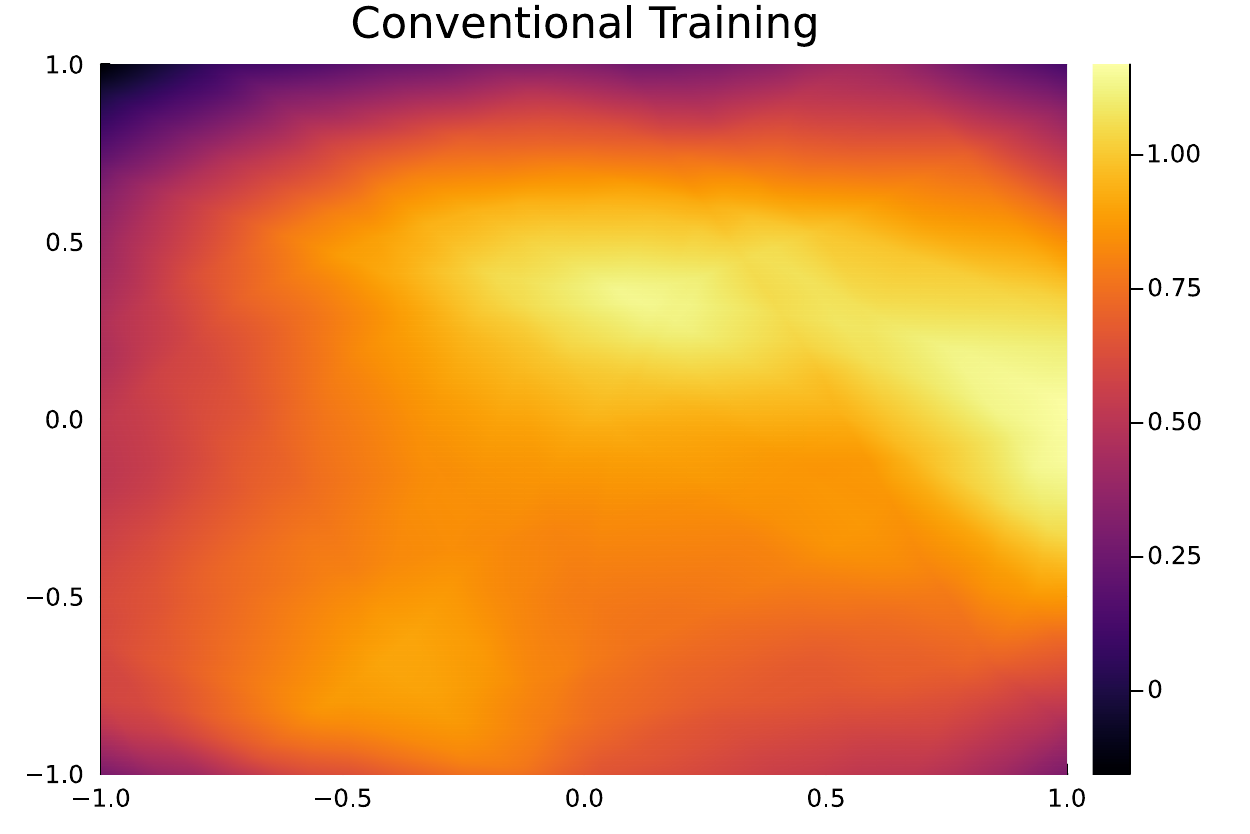}
            \caption{}
        \end{subfigure}
        \begin{subfigure}{0.49\textwidth}
            \includegraphics[width=.8\linewidth]{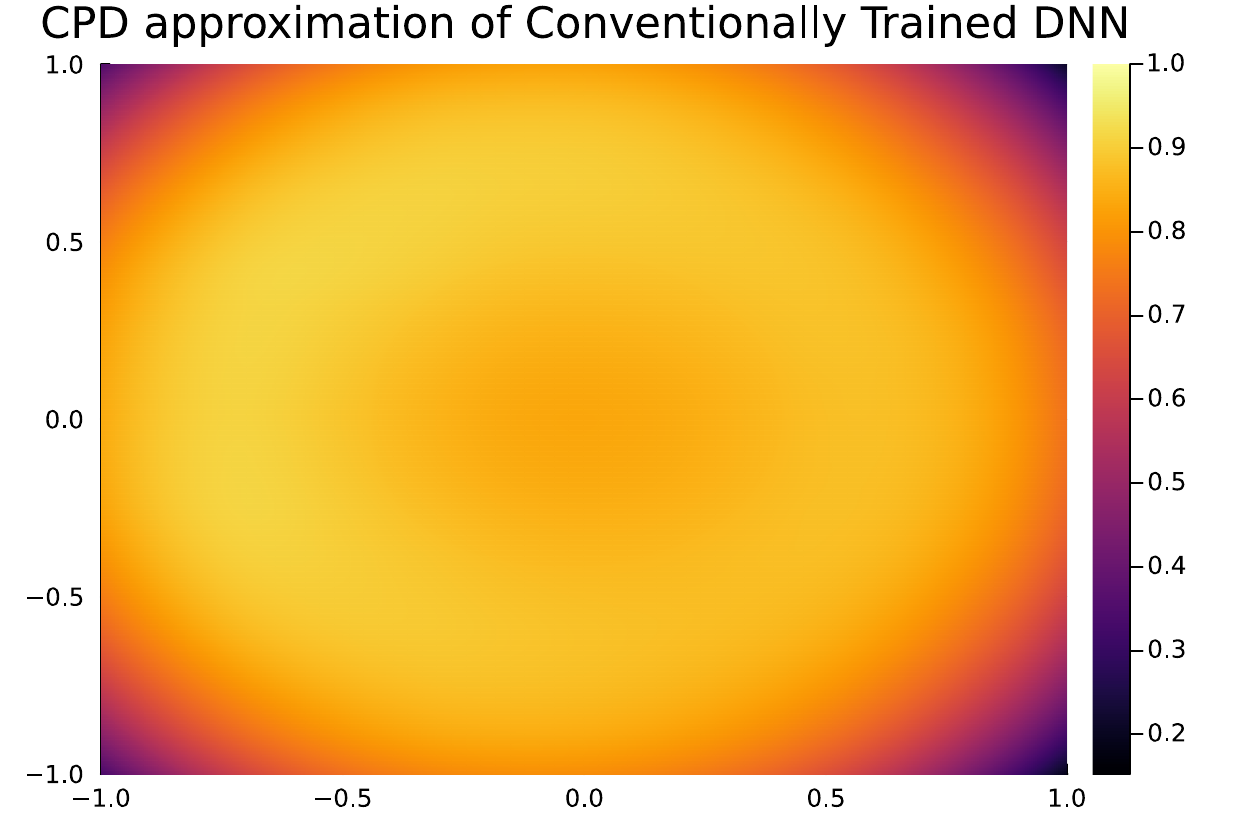}
            \caption{}
        \end{subfigure}\hfill
        \begin{subfigure}{0.49\textwidth}
            \includegraphics[width=.8\linewidth]{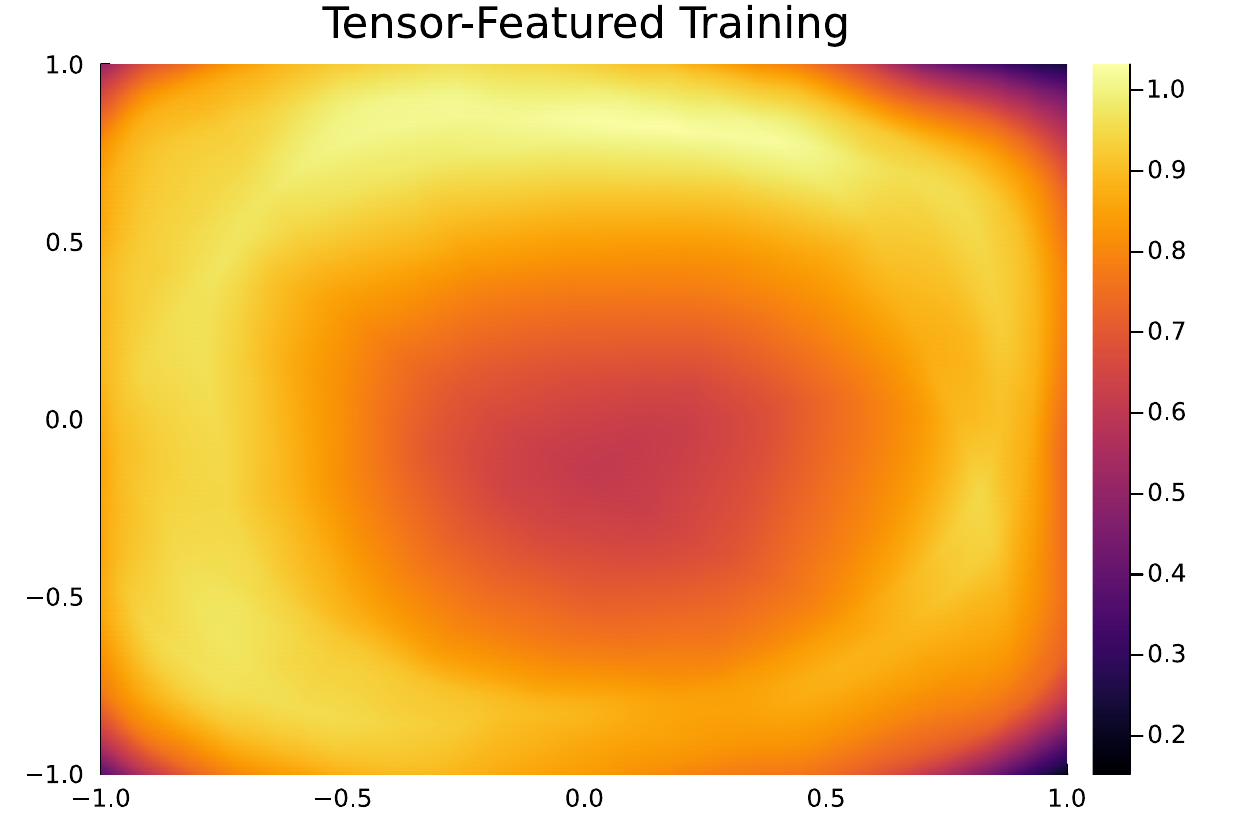}
            \caption{}
            \label{fig:tnheatmap}
        \end{subfigure}\hfill
        \caption{ (a) The convergence of different training procedures to learn \cref{eq:func1} with $d=15$.
        Heatmaps of (b) the function learned by the conventionally trained DNN, (c) the CPD approximation of the conventionally trained DNN and (d) the Tensor-Featured trained DNN. Heatmaps are evaluated on a 2000 point interpolation grid in all modes and all but 2 modes are fixed to the 500th point on their interpolation grid. The DNN training is fixed to 10,000 samples in the training set.}
        \label{fig:more_samples}
    \end{figure}
    
    In \cref{fig:more_samples}, we improve the accuracy of the conventionally trained DNN by increasing the training sample batch size from 1,000 to 10,000.
    With a qualitatively more accuracy reference, the CPD approximation generates a function that captures more character of the true function.
    Finally, equipped with this CPD, the Tensor-Featured training is greatly improved and the mean squared error in the validation set falls dramatically.
    In general, these results demonstrate that introduced features provide context for DNN training.
    This 2-step optimization process of training DNN with a fixed feature and decomposing partially trained DNN into CPD can be repeated iteratively in an attempt refine the Tensor-Featured DNN.
    
\begin{figure}
        \begin{subfigure}{0.49\textwidth}
            \includegraphics[width=.8\columnwidth]{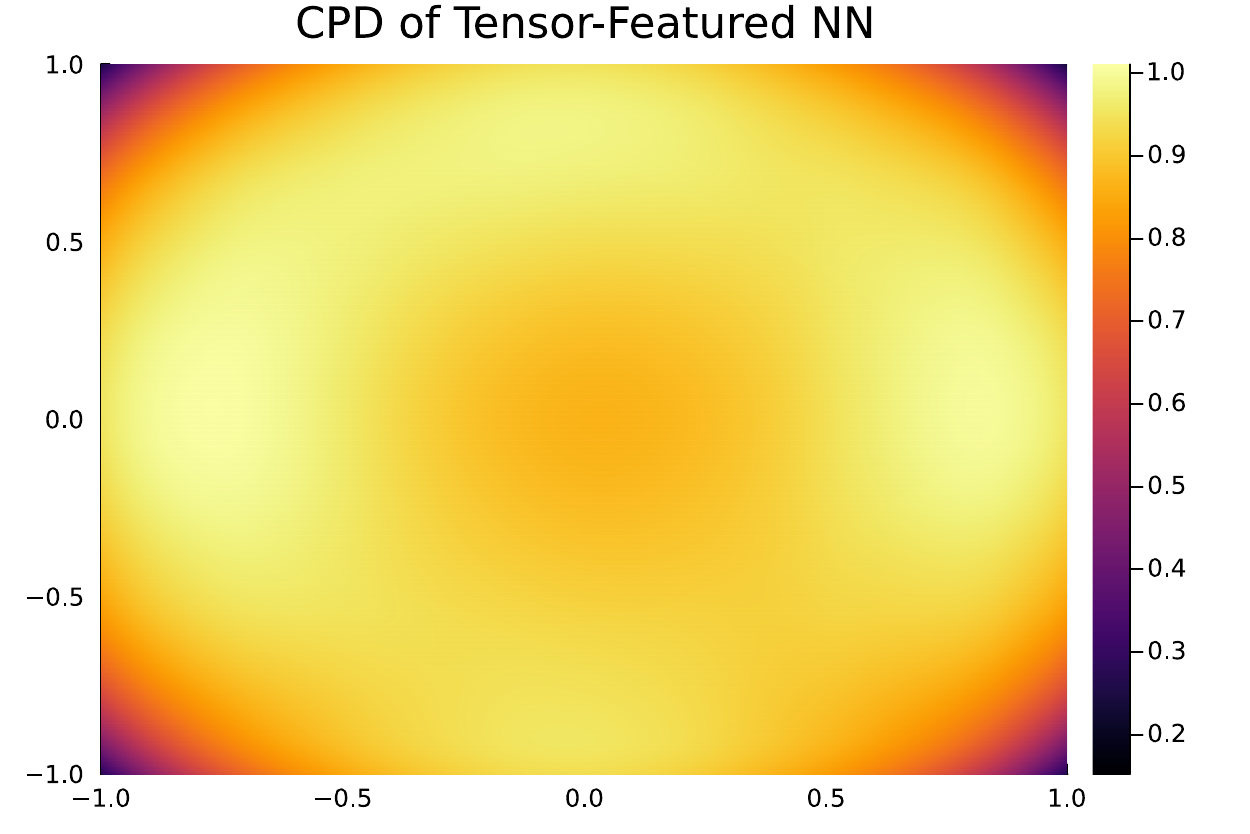}
            \caption{}
        \end{subfigure}\hfill
        \begin{subfigure}{0.49\textwidth}
            \includegraphics[width=.8\linewidth]{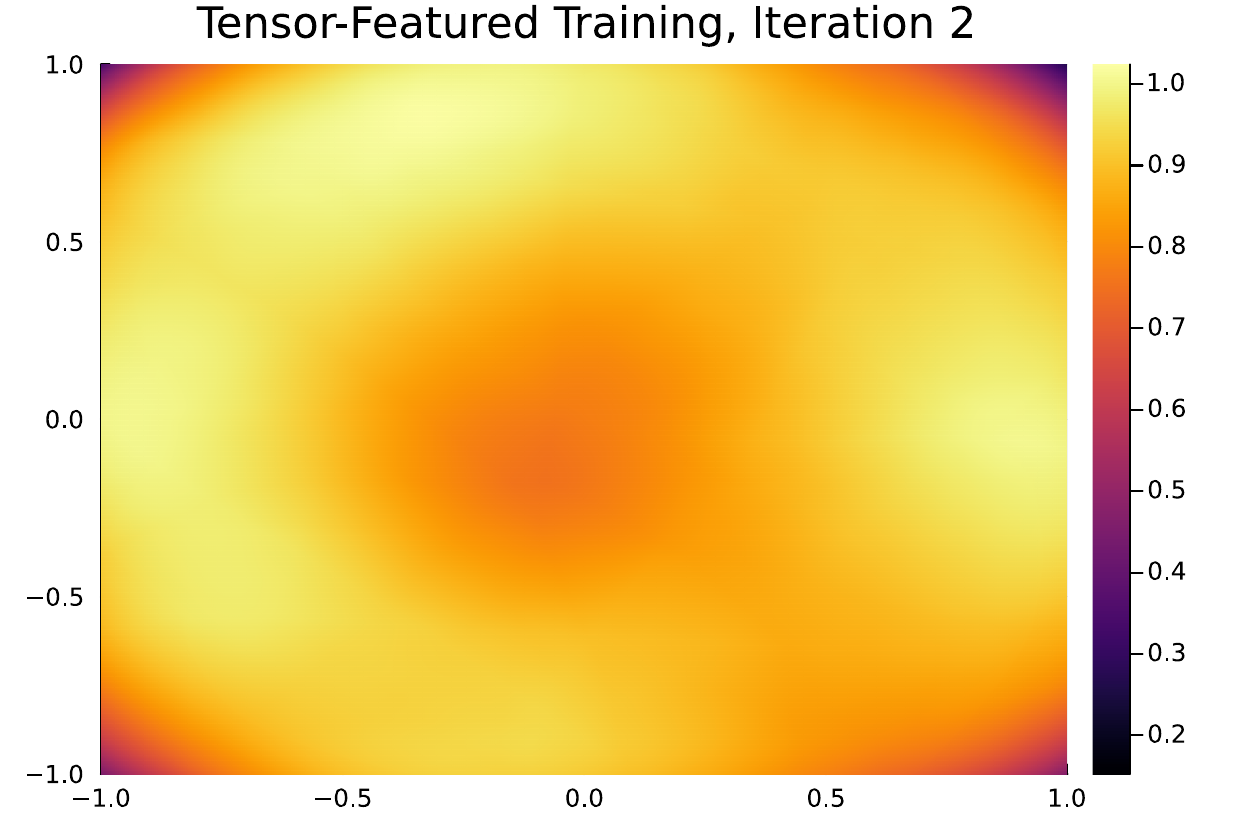}
            \caption{}
        \end{subfigure}
        \caption{Heatmaps of (a) the CPD approximation of the first Tensor-Featured trained DNN and (b) the second iteration Tensor-Featured trained DNN. Heatmaps are evaluated on a 2000 point interpolation grid in all modes and all but 2 modes are fixed to the 500th point on their interpolation grid. The DNN training is fixed to 10,000 samples in the training set.}
        \label{fig:second_iter}
    \end{figure}

    For example, we train the 15 dimensional example from above in a 3 step optimization procedure.
    First a DNN is trained conventionally using 10,000 samples and 3000 training epoch.
    After the training session, the DNN is discretized and decomposed into a CPD. 
    This CPD is passed as a feature into a new DNN which trained for 3000 iterations using the same 10,000 training points with the value of the CPD interpolation at each training point added to the feature list. 
    After this second training session, the resulting DNN is again decomposed into a CPD and this CPD is passed to the training of a third DNN.
    In \cref{fig:second_iter} we show the same 2D heatmaps for the CPD of the Tensor Featured trained DNN (i.e. \cref{fig:tnheatmap}) and the DNN constructed from the second round of Tensor-Featured training.
    While it is hard to tell if this network is an improvement on the first iteration, the data becomes clear when considering the relative L2 error and absolute mean squared error on the 2D slices in \cref{tab:l2_error}.
    From these results one can see that this multi-step training process, iteratively reduces the error in the trained DNNs. 
    
    \begin{table}
        \centering
        \begin{tabular}{|c|c|c|c|}
            \hline & Conventional & Tensor-Featured Iteration 1 & Tensor-Featured Iteration 2 \\ \hline
            Relative L2 Error & $0.272$ & $0.119$ & $0.0619$  \\\hline
            Absolute Mean Squared Error & $6.32 * 10^{-2}$ & $1.20 * 10^{-2}$ & $3.26*10^{-3}$ \\ \hline
        \end{tabular}
        \caption{Relative L2 and absolute mean squared error in DNN predicted functions on a 2D slice discretized by 2000 points in each dimension.}
        \label{tab:l2_error}
    \end{table}
    
    Because the tensor features provide context to the DNN optimization, we require that the initial training be relatively close (in some measure) to the global extrema.
    If the original network is not within epsilon of the solution, the Tensor-Featured training doesn't help dramatically, if at all.
    In the example above, we were able to improve the canonical training solution by increasing the number of batched training point samples.
    In general, it may not be possible to improve the accuracy of conventional training so easily, for example there may be a limitation on the total number of available samples.
    We are therefore interested in developing a strategy that introduces tensor features into the DNN training process from the start.
    So far, we have only considered decomposing partially trained DNN however practically, we are able to introduce any one or many features into the Tensor-Featured training process. 
    In the next section we will explore the idea of adding tensor features beyond TN decomposed DNN.
    
    \subsection{Leveraging Tensor-Featured Context to Accelerate Training}
    One of the biggest challenges of the Tensor-Featured training procedure as presented in the previous section is need for features to provide meaningful context to a DNN's training.
    If a target function is unknown and the convergence of conventional DNN training is slow then the Tensor-Featured training as presented thus far will not, in general, improve DNN optimization.
    Therefore, to improve optimization, we consider introducing alternative and additional contextual features.

    In this section we define two types of features.
    The first type of features are elementary functions which can be evaluated quickly for points in the domain of the unknown function.
    Practically these features could be discretized and decomposed using TN methods but, because the functions can be evaluated quickly, we choose to simply access elements of the functions in a matrix-free manner.
    The first class of features we will call \textbf{Rank-1 features}.
    The second type of feature are more complex functions which may be either slow to evaluate for points in the functions domain or otherwise limit the Tensor-Featured optimization. 
    For this second class of features, we will leverage tensor decomposition strategies to address these issues. 
    An example of this type of feature are the partially trained DNN from the previous section.
    We call this second class of features \textbf{TN features}.

    In this section we will attempt to learn a more difficult function, the solution to the initial boundary value problem of the hyperbolic (wave) equation 
    \begin{align}\label{eq:funct2}
        u(t,x) = (\exp(t^2) - 1) \sin(\frac \pi 2 (1 -|x|)^{2.5})
    \end{align}
    with $d = 15$.
    In this study, we only consider the following rank-1 features
    \begin{itemize}
        \item $y = \sum_i x_i$
        \item $y = \|x\|_2$
        \item $y = \sin(\|x\|_2)$
        \item $y = \cos(\|x\|_2)$
        \item $y = \sum_i \sin(x_i)$
        \item $y = \sum_i \cos(x_i))$
        \item $y = \exp(\|x\|_2)$
    \end{itemize}
    which map $\mathbb{R}^{n} \rightarrow \mathbb{R}$.
    Our decision to use these rank-1 features is relatively arbitrary.
    Any function is acceptable as long as it map $\mathbb{R}^{k} \rightarrow \mathbb{R}^{m}$
    where $k \leq n$ and $\mathbb{R}^{m}$ is the dimensionality of function being learned.
    In follow-up studies we will investigate methods to automatically choose a good set of Rank-1 features.
    Furthermore, it should be noted that we will only consider the CPD of a previously trained DNN as a TN feature though in future work other TN features will be investigated.     
    \begin{figure}
        \begin{subfigure}{0.49\textwidth}
            \includegraphics[width=.8\columnwidth]{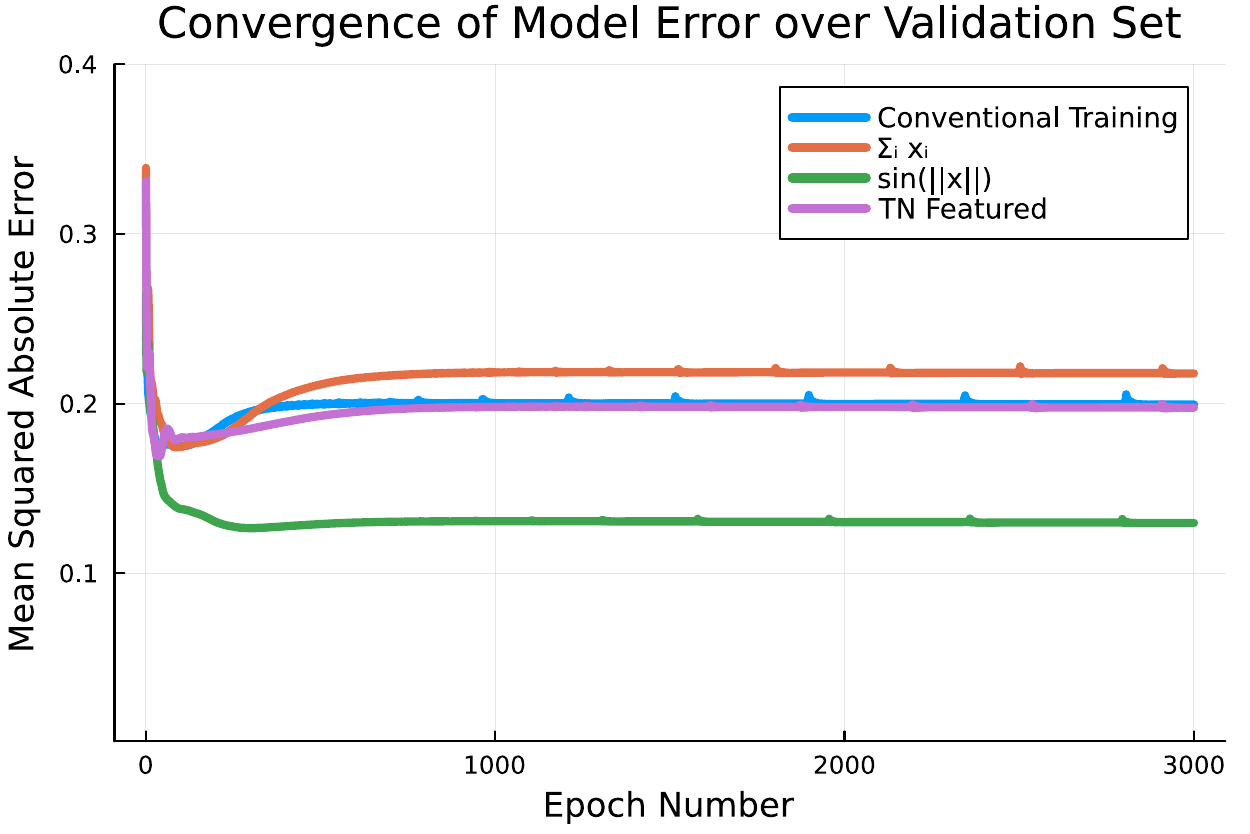}
            \caption{}
            \label{fig:1feature}
        \end{subfigure}\hfill
        \begin{subfigure}{0.49\textwidth}
            \includegraphics[width=.8\linewidth]{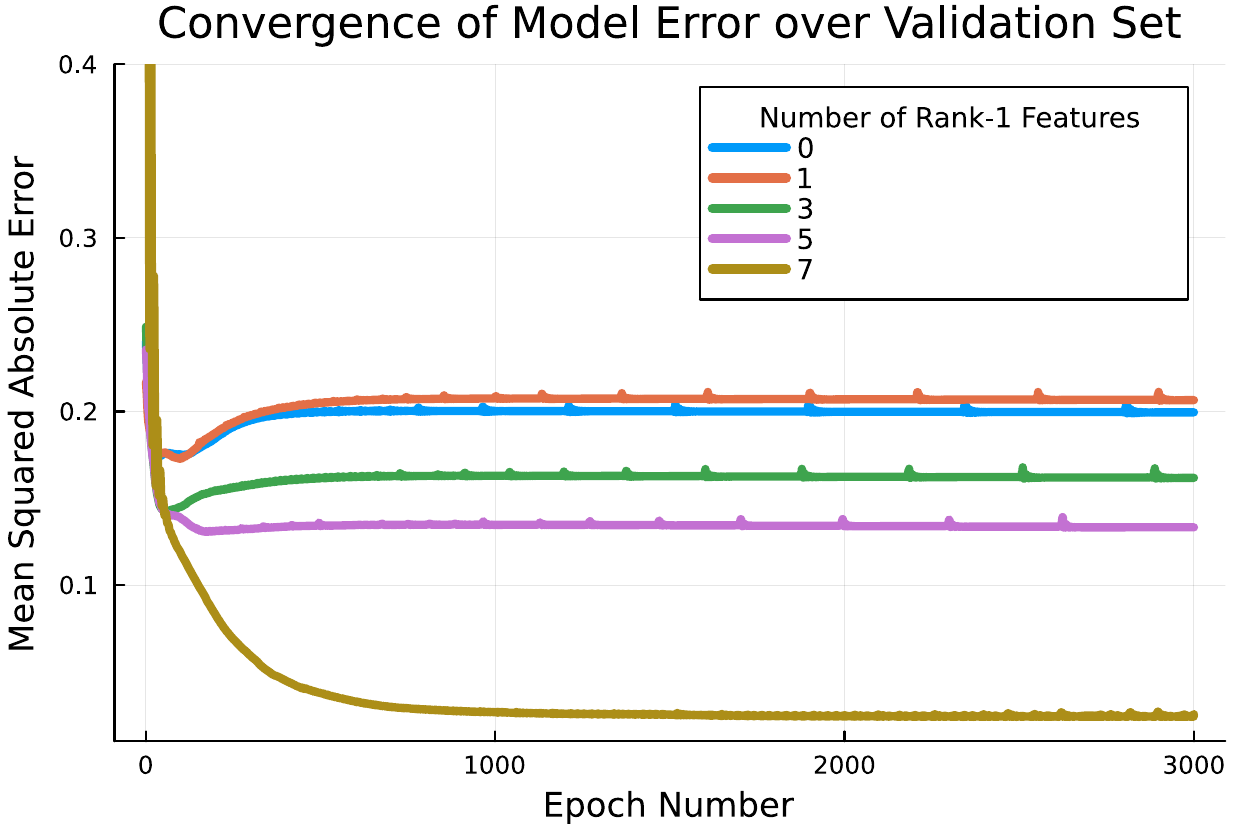}
            \caption{}
            \label{fig:multifeature}
        \end{subfigure}\hfill
        \centering
        \begin{subfigure}{0.49\textwidth}
            \includegraphics[width=.8\linewidth]{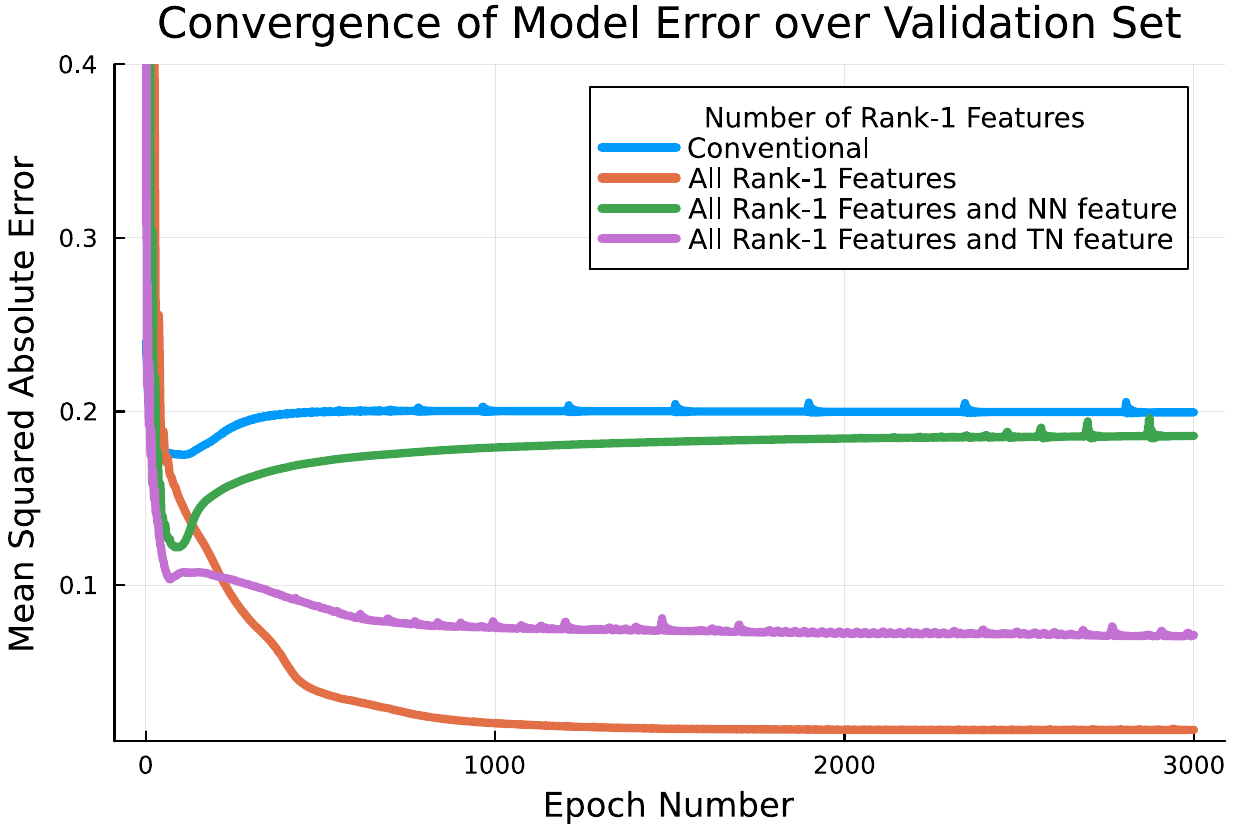}
            \caption{}
            \label{fig:poisonfeature}
        \end{subfigure}
        \caption{Comparison of the convergence of different training procedures to learn \cref{eq:funct2}. (a) Comparison of trainings with at most a single feature added to the DNN.
        (b) Comparison of trainings with single or multiple features added to the DNN.}
        \label{fig:feature-search}
    \end{figure}
    
    In \cref{fig:feature-search}, we consider the effect of using different features in the Tensor-Featured training.
    In \cref{fig:1feature} we add just a single to the training and find that the feature does influence the training efficiency.
    One can see that adding a feature that does not align with the problem does not significantly improve the DNN optimization performance.
    But adding a feature that does align with the problem improves the training.

    In \cref{fig:multifeature}, we consider the effect of adding multiple features in parallel to the Tensor-Featured training.
    In this figure, we incrementally introduce the features from the rank-1 feature list above in the order of the list.
    One can see that introducing multiple features seems to systematically improve the accuracy over the conventional training and the best single feature training in \cref{fig:1feature}.
    In our tests we found that most of the functions in the rank-1 feature list do not significantly improve the DNN training but together they outperform any single feature.
    However, we note that this process can fail when a feature is added that puts the DNN close to a local extrema.
    For example in \cref{fig:poisonfeature}, we add the conventionally trained DNN into feature list and find that the optimization quickly converge back to the Conventional Training solution.
    Though one can see that replacing the conventionally trained DNN with its CPD approximation can resolve this issue.

    \begin{figure}
        \begin{subfigure}{0.49\textwidth}
            \includegraphics[width=.8\columnwidth]{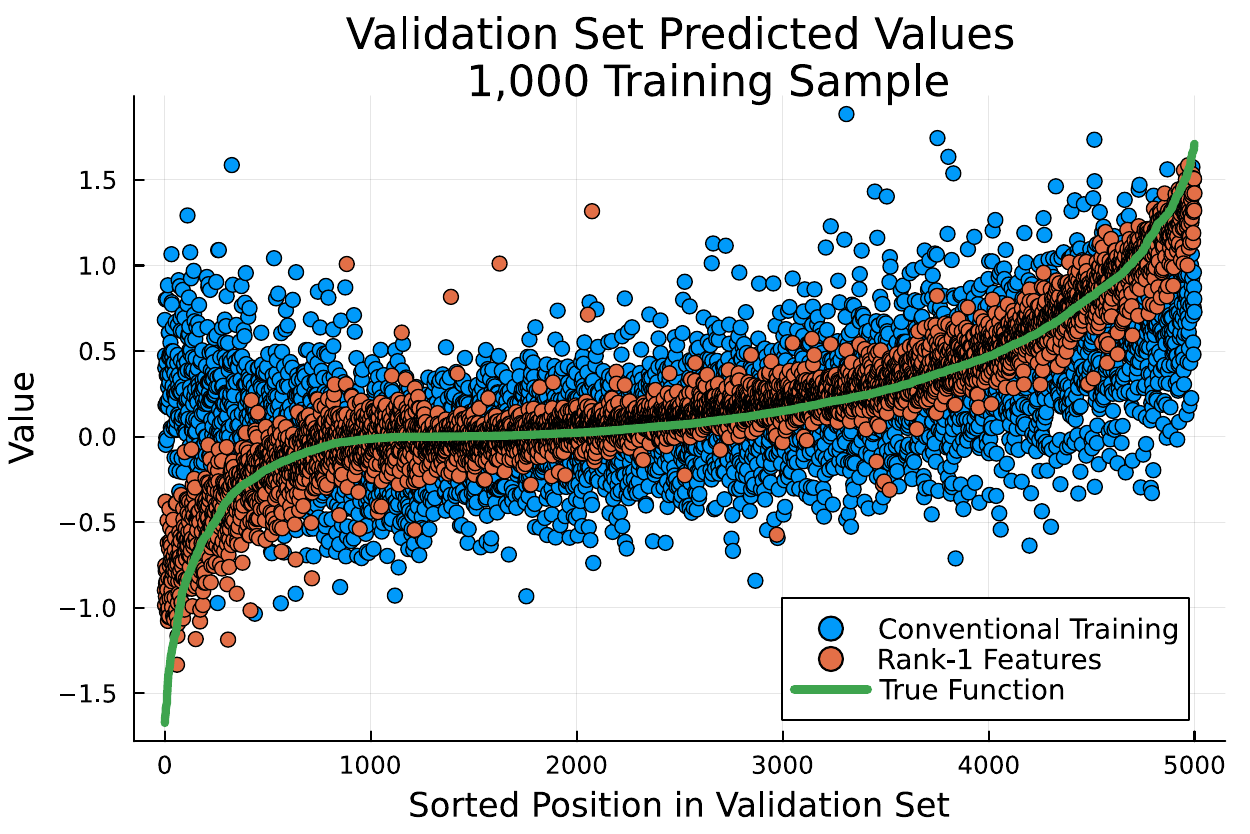}
            \caption{}
        \end{subfigure}\hfill
        \begin{subfigure}{0.49\textwidth}
            \includegraphics[width=.8\linewidth]{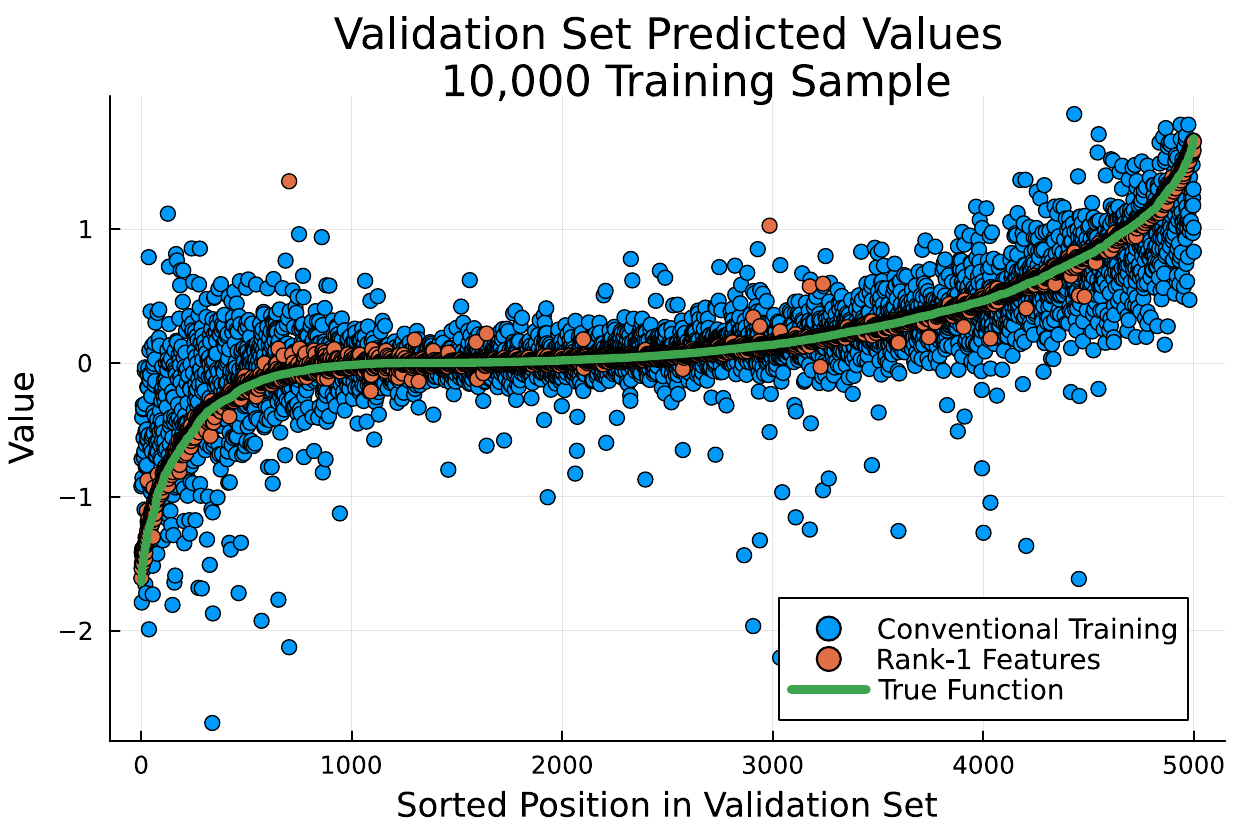}
            \caption{}
            \label{fig:15d_10000}
        \end{subfigure}\hfill
        \caption{Comparing the true values of \cref{eq:funct2} on the validation set 
        to those predicted by trained DNN sorted from most negative to most positive. (a) Fixes the training batch size to 1,000 points and (b) fixes the training batch size to 10,000 points.}
        \label{fig:scatter}
    \end{figure}
    Using Tensor-Featured training with rank-1 features, we can now improve the efficiency of training a DNN with no prior knowledge of the function.
    Furthermore, the rank-1 features do not depend on a pretrained DNN, like the CPD approximation did in the previous section.
    In \cref{fig:scatter}, we compare the accuracy of the conventional training to the Tensor-Featured training with only rank-1 features by plotting the true and predicted function values of points in the validation set.
    Note that the points are ordered via the value of the true function from most negative to most positive.
    One can see that the Tensor-Featured training using only rank-1 features more accurately predicts nearly all points in the validation set, compared to conventional training.
    Furthermore, the Tensor-Featured DNN training converges more quickly when the training batch size is increased.
    
\begin{figure}
        \begin{subfigure}{0.49\textwidth}
            \includegraphics[width=.8\columnwidth]{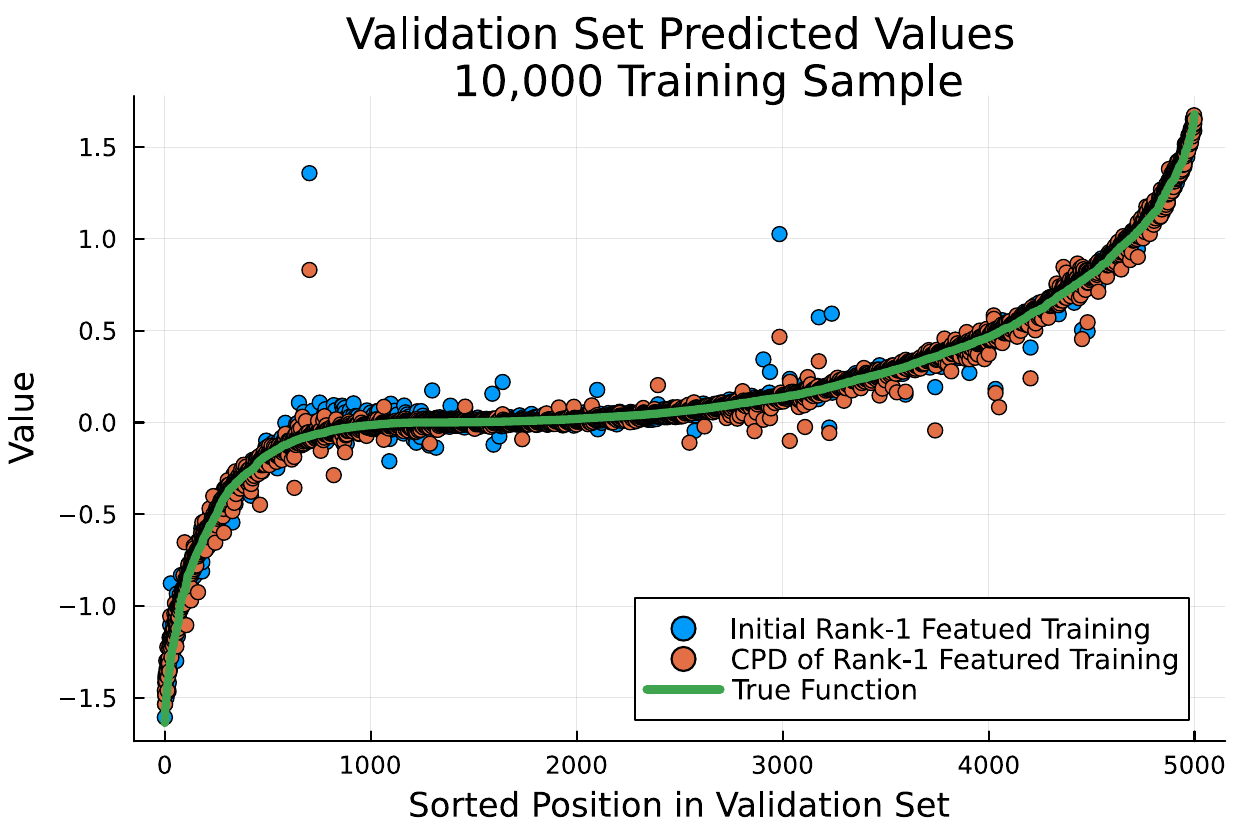}
            \caption{}
            \label{fig:scatterCPD}
        \end{subfigure}\hfill
        \begin{subfigure}{0.49\textwidth}
            \includegraphics[width=.8\linewidth]{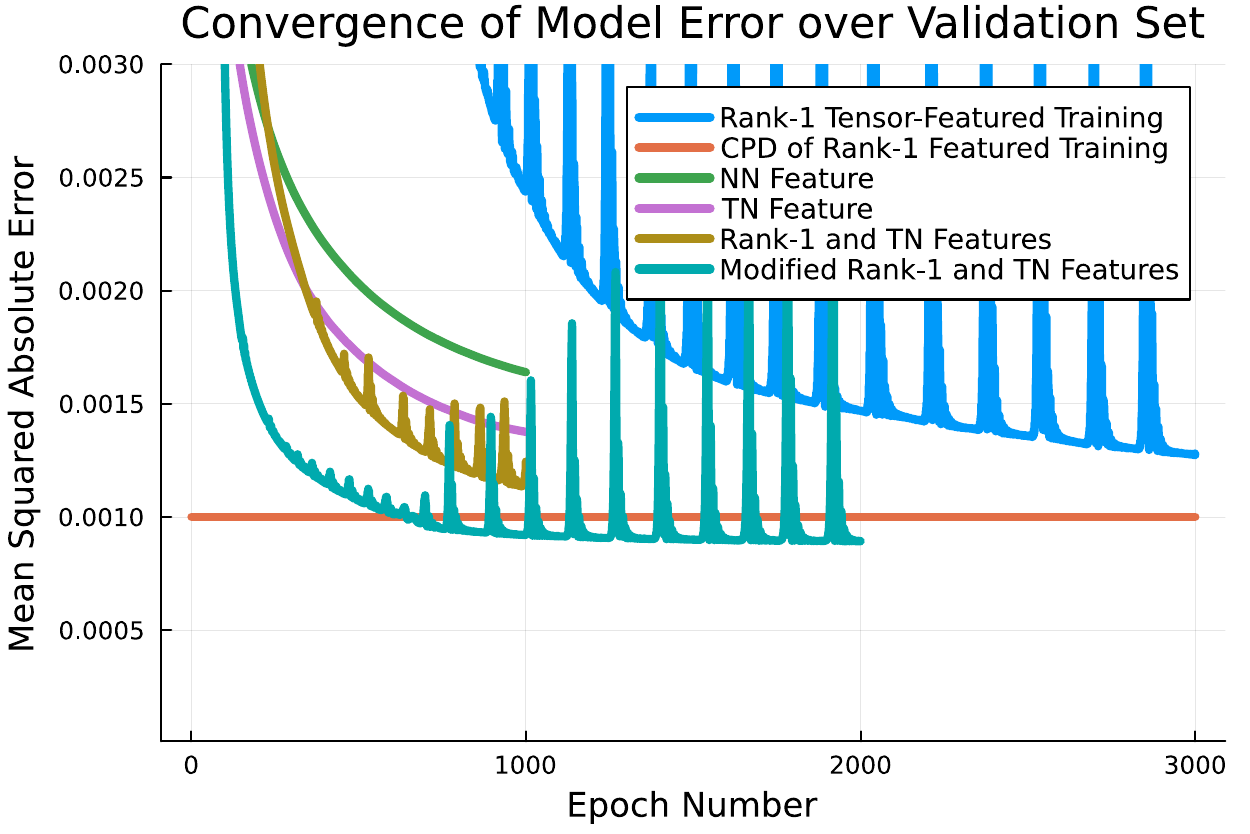}
            \caption{}
            \label{fig:scatter_iter2}
        \end{subfigure}\hfill
        \caption{
        (a) Comparing the true values of \cref{eq:funct2} on the validation set 
        to those predicted by a Tensor-Featured trained DNN and the CPD of the Tensor-Featured trained DNN sorted from most negative to most positive.
        (b) Comparison of the convergence of different training procedures to learn \cref{eq:funct2}.
        All trainings utilize a training batch size of 10,000 points.
        }
        \label{fig:scatter2}
    \end{figure}

    After training the DNN with rank-1 features, we can discretize and decompose the trained DNN as a TN.
    We compare the accuracy of the Tensor-Featured trained DNN to those interpolated by the CPD of this DNN in \cref{fig:scatterCPD}. 
    From this figure, one can see that the decomposition process has little impact on the accuracy of predicting values in the validation set. 
    Furthermore, by decomposing the DNN we, effectively, cannibalize all of the features from the preliminary DNN optimization into a single function.
    We demonstrate this in \cref{fig:scatter_iter2}
    where we attempt to improve on the first Tensor-Featured training.
    In the curve marked `NN Feature` we use exactly the DNN trained via the Rank-1 Tensor Featured Training into the optimization of a new DNN. 
    To do this optimization, we must evaluate the original DNN on the batch of training and validation points.
    This evaluation may be relatively slow, especially if there were many input features in this DNN or the DNN is very deep.
    The curve marked `TN Feature` uses only CPD interpolation of the Rank-1 Featured Training.
    Similar to the previous results, the CPD approximation of the DNN is more accurate than the DNN.
    For the curve marked `Rank-1 and TN Features` we combine the CPD with the original Rank-1 features and find that the new Tensor-Featured Training can become more accurate than the original Tensor-Featured Training.
    Because the original features were well captured by the CPD approximation, we are also free to modify the feature list in the DNN retraining. For example for the curve marked `Modified Rank-1 and TN Features` we remove features 1, 2, 5 and 6 from the input and find a result that outperforms the CPD approximation alone.
    In these optimization curves we often see oscillation in the errors and will investigate this phenomenon in future studies.
    This two step process of training DNN and updating features can be repeated iteratively until some convergence is reached. 

\begin{figure}
            \includegraphics[width=.7\columnwidth]{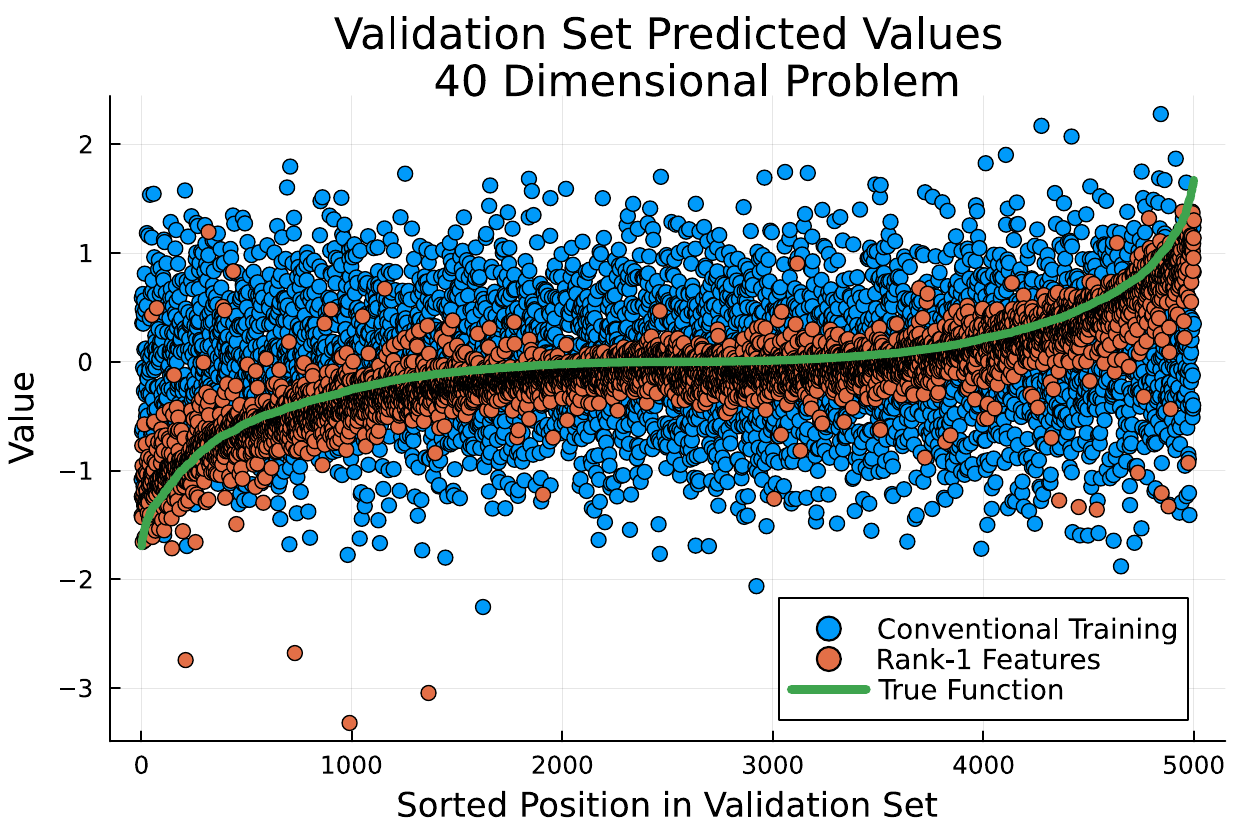}
        \caption{
        (a) Comparing the true values of \cref{eq:funct2} where $d=40$ on the validation set 
        to those predicted by a Tensor-Featured trained DNN sorted from most negative to most positive.
        Trainings utilize a training batch size of 10,000 points.
        }
        \label{fig:scatter40d}
    \end{figure}

    Finally in \cref{fig:scatter40d}, we the accuracy of training a DNN to learn \cref{eq:funct2} in 40 dimensions with either conventional training or Tensor-Featured training using only rank-1 features.
    One can see that the Tensor-Featured training more accurately learns the objective function in just 3000 epoch.
    Though comparing this figure to \cref{fig:15d_10000}, one can see that increasing the dimensionality of the problem slows the convergence of both training procedures. In \cref{tab:40dcost}, we show the cost comparing conventional training to Tensor-Featured with only rank-1 features and see that the Tensor-Featured training is only marginally more expensive in both optimization time and storage complexity.

    \begin{table}
        \centering
        \begin{tabular}{|c|c|c|c|}\hline
             & Time (s) & Parameters & Storage (KiB) \\ \hline
           Conventional Training  & 76.93 & 14,301 & 112.031 \\\hline
            Rank-1 Features & 77.47 & 15,001 & 117.500 \\ \hline
        \end{tabular}
        \caption{Costs associated with training different networks to represent \cref{eq:funct2} in 40 dimensions }
        \label{tab:40dcost}
    \end{table}

\section{Conclusion and Outlook}\label{sec:conclusion}
In this work we introduce an strategy to add context-oriented features into the input layer of a DNN and leverage tensor decomposition strategy to develop a 2-step optimization strategy for DNN training.
We are motivated by the idea of introducing pretrained DNN as input features similar to a warm-start initialization strategies and by the outer product structure of tensor decompositions.
We find that DNN optimized with a pretrained DNN feature quickly converge to the same local minima as their pretrained feature.
Therefore, to improve this optimization strategy, we discretize the DNN over the domain of the problem on a quadrature grid and decompose the resulting higher-order tensor using efficient randomized tensor decomposition strategies.
We restrict this work by only considering the CPD and to optimize the CPD we utilize a leveraged-score based column selected randomized ALS scheme.
This CPD-RALS strategy reducing the overall memory requirements of the decomposition by between 8 and 34 orders of magnitude.
We recognize that tensor decomposition algorithms like the ALS have a relationship to multidimensional gaussian process strategies.
This global optimization strategy appears to have a smoothing effect on functions that are encoded in trained  DNN.
We introduce this TN approximation as a feature in the DNN optimization as a means to efficiently kick optimizations out of local minima.

Our initial testing shows mixed results when the only feature introduced is the TN decomposition of a DNN.
Primarily, we found that, if the trained DNN is relatively close to a global extrema, Tensor-Featured training can significantly improve the optimization of the subsequent DNN.
However, if the trained DNN is not close to a global extrema then the tensor feature does not improve the training, i.e. the context provided by the tensor feature is irrelevant.
We recognize that it may be difficult to train a DNN to be sufficiently close to a global extrema, which limits the utility of the Tensor-Featured training as described above. 
As a results, we introduce a more robust strategy to improve optimization.

For this robust strategy, we recognize it is acceptable to introduce any number of functions that acts on any portion of the domain as features to the DNN optimization.
We therefore introduce an improved method where many functions, which we believe may provide context, are added as features to the DNN optimization. 
The category of features can be split into functions which are easy to apply to points on the domain, which we call rank-1 features, and functions which must first be decomposed as a TN to be applied efficiently, i.e. partially optimized DNN.
Using a combination of rank-1 and TN features, we are able to significantly accelerate the convergence of model error in the validation set compared to conventional DNN training procedures.

In follow-up work, we will consider methods to efficiently choose rank-1 features, determine the location and number of quadrature grid points for TN decomposition strategies and pick which TN decomposition strategy is necessary for particular problems.
We are also interested in studying how the product structure of these features influences the optimization of parameters in a DNN. 
For example, when does adding a feature improve optimization, how do different features influence optimization of DNN parameters, and is there an interplay between activation functions and rank-1 features?
Finally, in this work we found some difficulties when optimizing extremely high dimensional functions using the randomized CPD-ALS. 
We are, therefore, interested in determining methods to improve the CPD optimization such as leveraging the symmetry and sparsity of high-dimensional PDEs.

\bibliography{references}
\end{document}